\documentclass[twocolumn,11pt,a4paper]{article} 
\usepackage{booktabs}
\usepackage{multirow}
\usepackage[utf8]{inputenc}
\usepackage[T1]{fontenc}

\usepackage{amsmath,amssymb,mathtools}

\usepackage{graphicx}
\usepackage{subfig}
\usepackage{threeparttable}
\usepackage{float}

\usepackage[ruled,vlined]{algorithm2e}
\usepackage{algorithmic}

\usepackage[numbers]{natbib}  
\usepackage[hidelinks]{hyperref}
\usepackage{url}

\usepackage[margin=0.75in]{geometry}

\title{Adaptive Malware Detection using Sequential Feature Selection:\\ A Dueling Double Deep Q-Network (D3QN) Framework for Intelligent Classification}

\author{
Naseem Khan\thanks{Department of Computer Science and Engineering, Hamad Bin Khalifa University, Doha, Qatar. Email: nakh12498@hbku.edu.qa} \and
Aref Y. Al-Tamimi\thanks{Qatar Computing Research Institute, Hamad Bin Khalifa University, Doha, Qatar. Email: ikhalil@hbku.edu.qa} \and
Amine Bermak\footnotemark[1] \and
Issa M. Khalil\footnotemark[2]
}

\date{} 

\begin{document}

\maketitle

\begin{abstract}
Traditional malware detection methods exhibit computational inefficiency due to exhaustive feature extraction requirements, creating accuracy-efficiency trade-offs that limit real-time deployment. We formulate malware classification as a Markov Decision Process with episodic feature acquisition and propose a Dueling Double Deep Q-Network (D3QN) framework for adaptive sequential feature selection. The agent learns to dynamically select informative features per sample before terminating with classification decisions, optimizing both detection accuracy and computational cost through reinforcement learning.

We evaluate our approach on Microsoft Big2015 (9-class, 1,795 features) and BODMAS (binary, 2,381 features) datasets. D3QN achieves 99.22\% and 98.83\% accuracy respectively while utilizing only 61 and 56 features on average, representing 96.6\% and 97.6\% dimensionality reduction compared to full feature sets. This yields computational efficiency improvements of 30.1× and 42.5× over traditional ensemble methods. Comprehensive ablation studies demonstrate consistent superiority over Random Forest, XGBoost, and static feature selection approaches across all performance metrics.

Quantitative analysis demonstrates that D3QN learns non-random feature selection policies with 62.5\% deviation from uniform baseline distributions across feature categories. The learned policies exhibit structured hierarchical preferences, utilizing high-level metadata features for initial assessment while selectively incorporating detailed behavioral features based on classification uncertainty. Feature specialization analysis reveals 57.7\% of examined features demonstrate significant class-specific discrimination patterns.
Our results validate reinforcement learning-based sequential feature selection for malware classification, achieving superior accuracy with substantial computational reduction through learned adaptive policies rather than static dimensionality reduction techniques.
\end{abstract}


\section{Introduction}
Malware detection systems require real-time classification capabilities while maintaining high accuracy against evolving threats. Current supervised learning approaches exhibit a fundamental computational bottleneck: they extract complete feature sets from executable files regardless of sample complexity, creating uniform processing overhead that constrains deployment in resource-limited environments \cite{souri2018state, raff2020survey}. This exhaustive feature extraction paradigm becomes particularly problematic when processing large sample volumes or operating under strict latency constraints \cite{choudhary2020malware, abisoye2025using}.

Existing dimensionality reduction techniques attempt to address computational complexity through global feature selection methods such as Principal Component Analysis and mutual information ranking \cite{bai2014malware, suarez2017droidsieve, rafique2019malware}. However, these approaches optimize feature subsets based on aggregate training statistics, potentially discarding features with high discriminative power for specific malware families or obfuscated variants. The resulting static feature selections cannot adapt to individual sample characteristics, leading to suboptimal accuracy-efficiency trade-offs \cite{toan2022static}.

We address this limitation by reformulating malware classification as a sequential decision-making problem under reinforcement learning. Our approach trains a Dueling Double Deep Q-Network (D3QN) agent to learn adaptive feature selection policies that dynamically determine which features to examine for each sample before making classification decisions. Unlike conventional methods that process predetermined feature sets, the agent optimizes computational resource allocation based on sample-specific characteristics within a Markov Decision Process framework.

\begin{figure*}[ht]
    \includegraphics[width=0.8\linewidth]{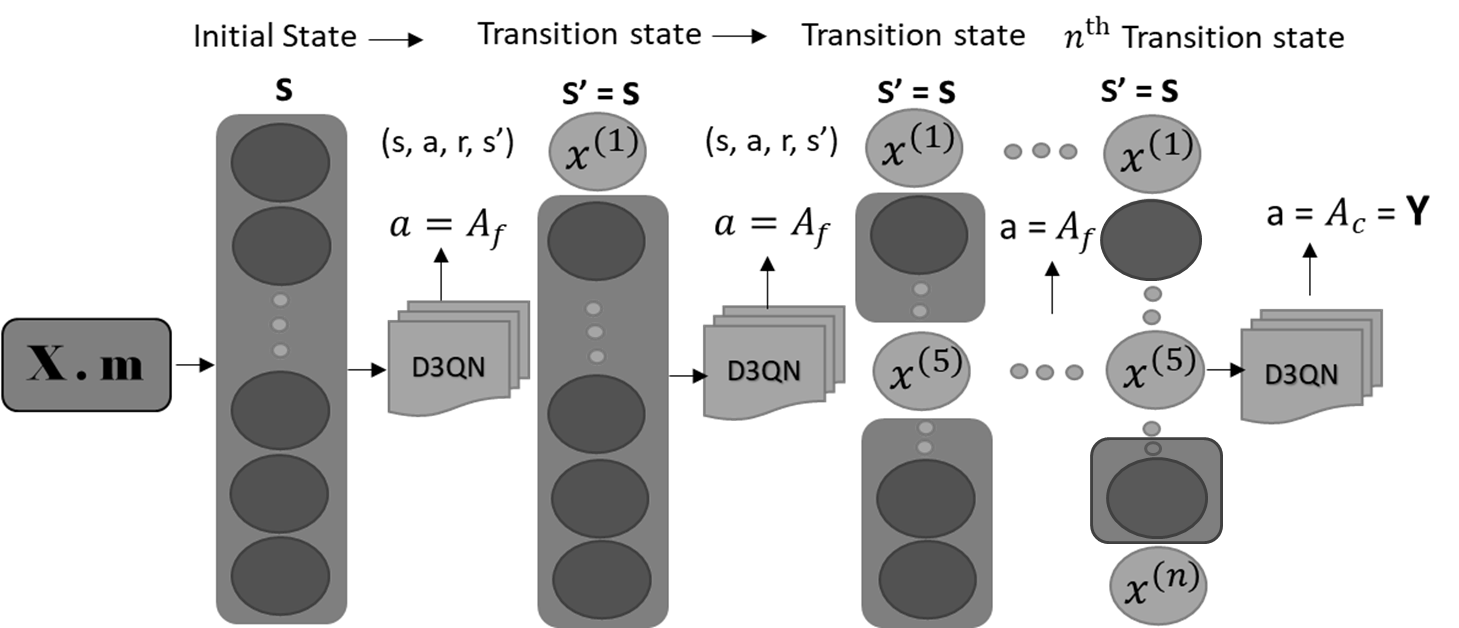}
    \centering
    \caption{Sequential decision-making framework for malware classification. The RL agent executes feature selection actions (\(A_f\)) to iteratively unmask attributes from an input sample (\(\textbf{X}\)), acquiring a subset of crucial features (\(f \subseteq F\)). Upon convergence to an optimal representation, the agent invokes a classification action (\(A_c\)) to map \(X\) to its class label (\(Y\)), maximizing cumulative reward under a D3QN policy for computational efficiency.}
    \label{fig:Decision-making-system}
\end{figure*}

The key innovation lies in episodic feature acquisition: the agent iteratively selects informative features or terminates with classification based on accumulated evidence, enabling sample-adaptive computational allocation. Simple samples are resolved with minimal features while complex cases justify additional analysis. Figure \ref{fig:Decision-making-system} presents the sequential decision-making framework. The D3QN architecture separates state value estimation from action advantage computation, facilitating effective learning in high-dimensional malware feature spaces.

Experimental validation on Microsoft Big2015 and BODMAS datasets demonstrates that our method achieves 99.22\% and 98.83\% accuracy while utilizing only 3.4\% and 2.4\% of available features respectively, yielding 30× computational efficiency improvements over ensemble methods. We introduce quantitative intelligence assessment metrics that provide empirical evidence of strategic learning behavior, showing the agent develops domain-specific feature hierarchies without explicit supervision. Comprehensive evaluation includes systematic comparison with traditional machine learning baselines and ablation studies demonstrating consistent superiority across multi-class and binary classification scenarios. Code and data are available at \url{https://github.com/Magnet200/D3QN.git}

This work establishes reinforcement learning as an effective paradigm for adaptive malware detection, enabling simultaneous optimization of accuracy and computational efficiency through learned sequential decision-making policies.

\section{Related Work}
\subsection{Deep Learning for Malware Detection}

Contemporary malware detection employs sophisticated deep learning architectures to improve classification performance. Recent advances include hybrid models combining static and dynamic analysis: Alsumaidaee et al. \cite{alsumaidaee2025optimizing} developed Static-CNN-LSTM and Dynamic-1D-CNN-LSTM architectures for real-time Android detection, while Yapici et al. \cite{yapici2025novel} employed ResNet, DenseNet, and ResNeXt on bytecode image representations. However, these approaches require exhaustive feature extraction, creating computational bottlenecks for real-time deployment. Gibert et al. \cite{gibert2025assessing} highlight additional robustness challenges against packing techniques, emphasizing the need for adaptive detection strategies.

\subsection{Feature Selection in Malware Detection}

Traditional feature selection applies global optimization techniques including PCA, Information Gain ranking, and tree-based importance measures \cite{bai2014malware, suarez2017droidsieve, rafique2019malware}. Recent work addresses specialized contexts: Panja et al. \cite{panja2025efficient} employ Extra Tree Classifier with Gini impurity for resource-constrained IoT environments, while Hasan et al. \cite{hasan2025enhancing} utilize LDA and PCA across multiple models for high-dimensional data management. These methods optimize feature subsets globally across training datasets, potentially discarding sample-specific discriminative information critical for variant detection.

\subsection{Reinforcement Learning for Sequential Decision-Making}

RL applications in malware detection focus primarily on sequential feature selection. Fang et al. \cite{fang2019feature} introduced DQFSA using Deep Q-Learning for PE feature selection, while Wu et al. \cite{wu2023droidrl} developed DroidRL with DDQN for Android malware detection. Both approaches achieve reduced feature sets but maintain architectural separation between feature selection and classification modules, limiting integrated optimization. Broader cybersecurity applications include adversarial training \cite{caminero2019adversarial} and network security \cite{al2023malbot}, though unified feature selection and classification remains underexplored.

\subsection{Research Gap and Positioning}

Existing methods exhibit two fundamental limitations: static global feature selection that ignores sample-specific characteristics, and decoupled architectures that prevent joint optimization of feature acquisition and classification decisions. Our D3QN framework addresses these gaps through unified episodic learning where a single agent learns both adaptive feature selection and optimal termination policies, enabling sample-specific computational allocation while maintaining classification accuracy.

\section{Methodology}

\subsection{Markov Decision Process Formulation}

We formulate malware classification as a sequential decision-making problem where an agent learns to adaptively select informative features before making classification decisions. The problem is modeled as a Markov Decision Process $(\mathcal{S}, \mathcal{A}, \mathcal{T}, \mathcal{R}, \gamma)$.

\textbf{State Representation}: Each state $s \in \mathbb{R}^{2n}$ concatenates the feature vector with a binary selection mask:
\begin{align}
s = [x; m] \in \mathbb{R}^{2n}
\end{align}
where $x = [x_1, \ldots, x_n]$ represents the complete feature vector of the malware sample, and $m = [m_1, \ldots, m_n]$ serves as a selection mask where $m_i \in \{0,1\}$ indicates whether feature $i$ has been revealed ($m_i = 1$) or remains unexplored ($m_i = 0$). Initially, $m = \mathbf{0}$ represents a completely unexplored sample where no features have been examined, simulating the scenario where the agent must sequentially decide which features to investigate.

\textbf{Action Space}: The action space comprises $|\mathcal{A}| = n + k$ total actions:
\begin{align}
\mathcal{A} = A_f \cup A_c
\end{align}
where $A_f = \{f_1, \ldots, f_n\}$ contains $n$ feature selection actions (corresponding to the $n$ features in the dataset), and $A_c = \{c_1, \ldots, c_k\}$ contains $k$ classification actions (corresponding to the $k$ malware classes). For our datasets, $n = 1795$ (Big2015) or $n = 2381$ (BODMAS), and $k = 9$ or $k = 2$ respectively.

\textbf{Transition Dynamics}: The deterministic transition function models two distinct behaviors as illustrated in Figure~\ref{fig:Decision-making-system}:
\begin{align}
\mathcal{T}(s, a) = \begin{cases}
s' \text{ with } m_i = 1 & \text{if } a = f_i \in A_f \\
\text{Terminal} & \text{if } a \in A_c
\end{cases}
\end{align}
Feature selection actions ($a \in A_f$) drive the agent to explore additional features by updating the corresponding mask element, enabling sequential feature revelation. Classification actions ($a \in A_c$) terminate the episode when the agent determines it has acquired sufficient features to make a confident prediction about the sample's class.

\textbf{Reward Structure}: The reward function balances classification accuracy with computational efficiency:
\begin{align}
\mathcal{R}(s, a) = \begin{cases}
-\lambda & \text{if } a \in A_f \\
\mathbb{I}[a = y] - 1 & \text{if } a \in A_c
\end{cases}
\end{align}
where $\lambda = 0.0001$ represents the feature acquisition cost, empirically determined through systematic evaluation of values in the range $[0, 1]$ to achieve optimal accuracy-efficiency trade-off. The indicator function $\mathbb{I}[\cdot]$ yields 0 for correct classification and -1 for misclassification, encouraging accurate predictions while penalizing feature usage.

\textbf{Action Masking Mechanism}: To ensure valid sequential behavior, we prevent selection of already-examined features through dynamic masking:
\begin{align}
\mathcal{M}(s, a) &= \begin{cases}
1 & \text{if } a = f_i \text{ and } m_i = 1 \\
0 & \text{otherwise}
\end{cases} \\
Q_{\text{valid}}(s, a) &= Q(s, a) - \mathcal{M}(s, a) \cdot 10^6
\end{align}
This mechanism assigns prohibitively low Q-values to invalid actions, ensuring the agent only selects previously unexplored features or proceeds to classification.

\subsection{Sample-Adaptive Learning Mechanism}

\textbf{Core Innovation}: Our approach enables adaptive computational resource allocation by learning sample-specific feature selection patterns. During training, the epsilon-greedy exploration strategy allows the agent to discover which features provide the most discriminative information for different sample types, ultimately learning to start feature exploration from globally important features and proceeding sequentially based on sample-specific characteristics.

\textbf{Adaptive Learning Process}: The agent employs epsilon-greedy exploration with linear decay:
\begin{align}
\epsilon_t = \max(0.03, 0.70 - \alpha \cdot t)
\end{align}

During the exploration phase, the agent randomly selects features (by setting $m_i = 1$ for various features) across diverse malware samples, learning feature discriminative patterns. Through this process, the agent develops expertise in identifying globally important features that provide initial classification signals, followed by sample-specific features that refine the decision. After training, the learned policy exhibits sample-adaptive behavior: for simple samples, initial features provide sufficient evidence for classification termination, while complex samples require additional feature exploration until confident classification becomes possible.

\textbf{Sequential Pattern Learning}: The epsilon-greedy mechanism enables the agent to learn internal sample patterns where the value of the first selected feature influences subsequent feature selection decisions. This creates a decision tree-like exploration where each revealed feature value guides the selection of the next most informative feature, ultimately leading to sample-specific termination when accumulated evidence indicates confident classification.

\begin{figure*}[htb]
\includegraphics[width=0.6\linewidth]{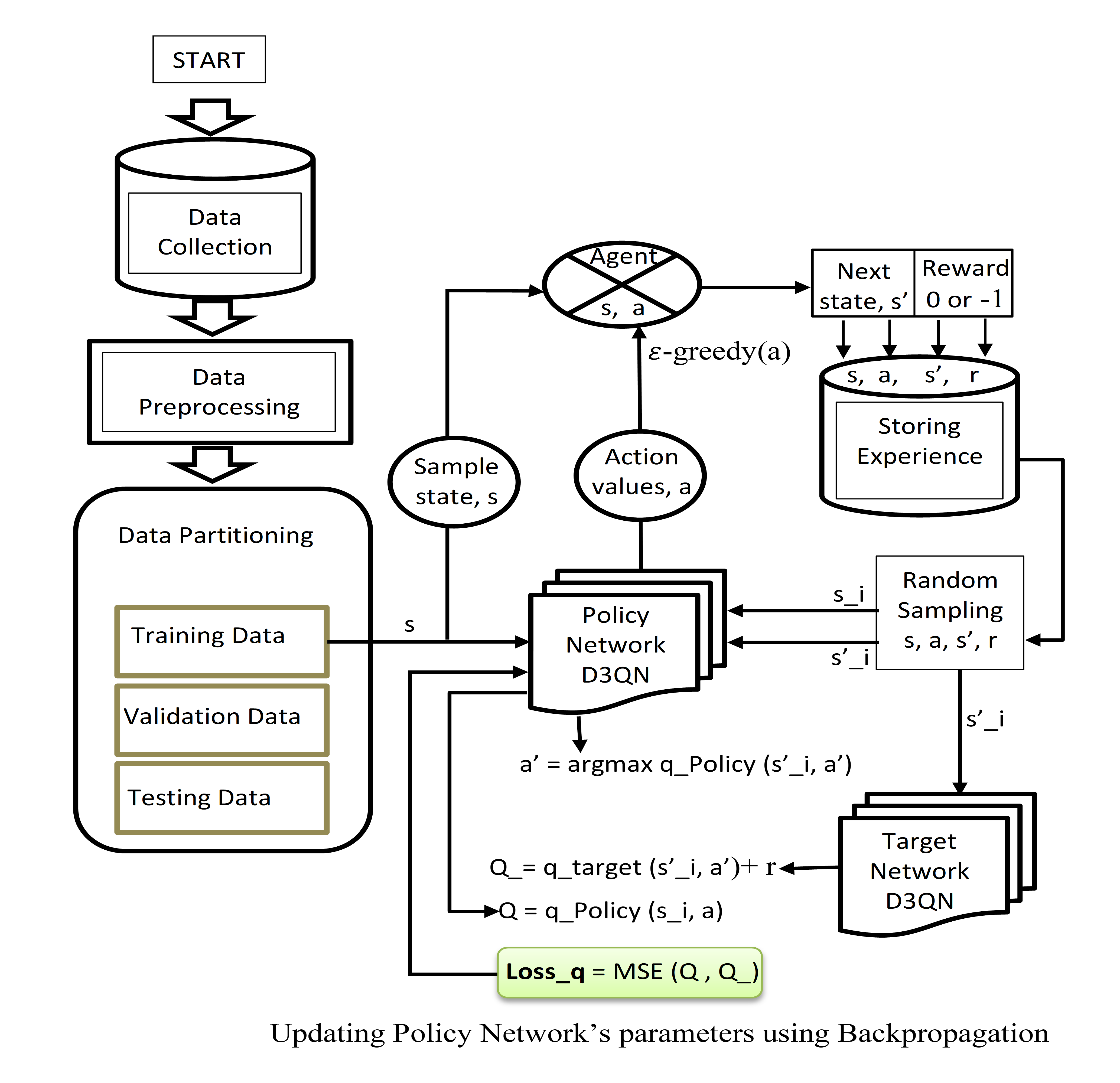}
\centering
\caption{Experimental flow-chart of our proposed policy network. Random sample, selected from training data as initial state $s$, passed through the policy network to approximate all the possible action values and from that values an optimal action value $a$ is selected by following $\epsilon$-greedy policy. By executing that action against the presented state, the agent get reward accordingly and the state transition which are stored as experiences. During training time, random experiences are sampled from the Replay-memory to optimise the policy network's parameters.}
\label{fig:D3QN_Model_archetecture}
\end{figure*}

\subsection{Dueling Double Deep Q-Network Architecture}

\textbf{D3QN Innovation}: Our proposed architecture decomposes Q-values to enable independent learning of state evaluation and action selection:

\textit{Shared Feature Extraction}: Three fully connected layers with PReLU activation process the concatenated state representation:
\begin{align}
h^{(1)} &= \text{PReLU}(W^{(1)} s + b^{(1)}) \\
h^{(2)} &= \text{PReLU}(W^{(2)} h^{(1)} + b^{(2)}) \\
h^{(3)} &= \text{PReLU}(W^{(3)} h^{(2)} + b^{(3)})
\end{align}

\textit{Dueling Stream Decomposition}: The shared representation branches into value and advantage estimation:
\begin{align}
V(s) &= W_v h^{(3)} + b_v \\
A(s, a) &= W_a h^{(3)} + b_a
\end{align}

\textit{Q-Value Aggregation with Mean Normalization}:
\begin{align}
Q(s, a) = V(s) + \Big(A(s, a) - \frac{1}{|\mathcal{A}|}\sum_{a'} A(s, a')\Big)
\end{align}

\textbf{Architectural Rationale}: The dueling decomposition enables the network to independently learn state value (how informative the current feature combination is) versus action advantages (which specific action provides the greatest benefit). This separation is crucial for sample-adaptive behavior as it allows the agent to recognize when sufficient evidence exists for classification termination regardless of which specific features remain unexplored.

\textbf{DDQN Baseline Architecture}: For comparative evaluation, we implement standard DDQN using identical network capacity with PReLU activation but direct Q-value estimation:
\begin{align}
Q(s,a) = W_{\text{out}} h^{(3)} + b_{\text{out}}
\end{align}
This baseline lacks the value-advantage decomposition, limiting its ability to distinguish termination timing from action selection.

\subsection{Training Algorithm and Implementation}

\textbf{Double Q-Learning Integration}: We employ double Q-learning to address overestimation bias inherent in standard Q-learning. The technique decouples action selection from value estimation: the online network selects the best action while the target network evaluates its value, reducing positive bias in Q-value updates:
\begin{align}
a^* &= \arg\max_{a'} Q_{\text{valid}}(s', a'; \theta) \\
Y_t &= r + \gamma Q(s', a^*; \theta^-)
\end{align}

\textbf{Soft Target Network Updates}: Rather than periodic hard updates, we employ exponential moving averages for target network parameter updates, providing training stability by gradually incorporating new learning while maintaining consistent target values:
\begin{align}
\theta^- \leftarrow \tau \theta + (1-\tau) \theta^-
\end{align}
where $\tau = 0.01$ ensures stable learning progression.

The complete training procedure is detailed in Algorithm~\ref{alg:D3QN_Training}, which integrates experience replay, action masking, and adaptive exploration within the reinforcement learning framework illustrated in Figure~\ref{fig:D3QN_Model_archetecture}.

\begin{algorithm}[ht]
\caption{D3QN Training for Sequential Feature Selection}
\label{alg:D3QN_Training}
\begin{algorithmic}[1]
\REQUIRE Dataset $\mathcal{D}$, training episodes $E$
\ENSURE Trained D3QN parameters $\theta$
\STATE Initialize D3QN with random $\theta$, $\theta^- \leftarrow \theta$
\STATE Initialize replay buffer $\mathcal{B}$, set $\epsilon \leftarrow 0.70$
\FOR{episode $e = 1$ to $E$}
    \STATE Sample $(x, y) \sim \mathcal{D}$, initialize $s \leftarrow [x; \mathbf{0}]$
    \WHILE{episode active}
        \STATE Compute masked Q-values: $Q_{\text{valid}}(s, \cdot)$
        \STATE Select action: $a \sim \epsilon\text{-greedy}(Q_{\text{valid}}(s, \cdot))$
        \IF{$a = f_i \in A_f$}
            \STATE Update mask: $m_i \leftarrow 1$, \\ observe reward $r = -\lambda$
        \ELSE
            \STATE Terminate episode, observe $r = \mathbb{I}[a = y] - 1$
        \ENDIF
        \STATE Store transition $(s, a, r, s')$ in $\mathcal{B}$
    \ENDWHILE
    \STATE Sample batch from $\mathcal{B}$, compute targets via double Q-learning
    \STATE Update $\theta$ via gradient descent on temporal difference error
    \STATE Soft update: $\theta^- \leftarrow \tau\theta + (1-\tau)\theta^-$
    \STATE Linear decay: $\epsilon \leftarrow \max(0.03, \epsilon - \alpha)$
\ENDFOR
\end{algorithmic}
\end{algorithm}

\textbf{Hyperparameter Configuration}: Complete training parameters including learning rates, batch sizes, and network architectures are provided in the supplementary materials (Table \ref{tab:dataset_baseline}). Key parameters were selected through systematic validation to ensure reproducible results across both datasets.

\textbf{Convergence Analysis}: The training dynamics create sample-adaptive behavior through the interaction of epsilon decay and reward structure. Initially, high exploration enables the agent to discover feature importance patterns across diverse samples. As epsilon decreases, the agent exploits learned patterns, developing policies that terminate early for samples where initial features provide sufficient classification confidence, while continuing exploration for ambiguous samples requiring additional evidence. This learning mechanism directly produces the observed computational efficiency gains where simple samples utilize fewer features than complex samples, achieving automatic resource allocation without predetermined stopping criteria.

\section{Experimental Setup and Evaluation}

\begin{table*}[htbp]
\centering
\caption{Comprehensive Performance Comparison of Malware Detection Approaches Across Multi-class and Binary-class Datasets}
\label{tab:main_comparison}
\resizebox{\textwidth}{!}{%
\begin{tabular}{lcccccccc}
\toprule
\multirow{2}{*}{\textbf{Method}} & \multicolumn{4}{c}{\textbf{Big2015 Dataset (9 classes, 1795 features)}} & \multicolumn{4}{c}{\textbf{BODMAS Dataset (2 classes, 2381 features)}} \\
\cmidrule(lr){2-5} \cmidrule(lr){6-9}
& \textbf{Accuracy (\%)} & \textbf{Precision (\%)} & \textbf{Recall (\%)} & \textbf{F1-Score (\%)} & \textbf{Accuracy (\%)} & \textbf{Precision (\%)} & \textbf{Recall (\%)} & \textbf{F1-Score (\%)} \\
\midrule
\multicolumn{9}{c}{\textit{Traditional Machine Learning Methods (All Features)}} \\
\midrule
Decision Tree & 94.25 & 94.09 & 94.25 & 94.11 & 95.76 & 95.76 & 95.76 & 95.76 \\
Random Forest & 98.02 & 97.68 & 98.02 & 97.84 & 98.10 & 98.10 & 98.10 & 98.10 \\
Logistic Regression & 97.65 & 97.96 & 97.65 & 97.74 & 98.19 & 98.21 & 98.19 & 98.20 \\

\midrule
\multicolumn{9}{c}{\textit{Ensemble Methods (All Features)}} \\
\midrule
XGBoost & 98.85 & 98.87 & 98.85 & 98.85 & 98.11 & 98.12 & 98.11 & 98.11 \\
Voting Classifier & 98.11 & 98.16 & 98.11 & 98.11 & 98.50 & 98.50 & 98.50 & 98.50 \\
Stacking Classifier & 97.93 & 98.62 & 97.93 & 98.19 & 98.60 & 98.60 & 98.60 & 98.60 \\
\midrule

\midrule
\multicolumn{9}{c}{\textit{Proposed Reinforcement Learning Methods with dynamic feature reduction}} \\
\midrule
DDQN (Baseline) & \underline{99.12} & \underline{99.14} & \underline{99.13} & \underline{99.13} & \underline{98.83} & \underline{98.49} & \textbf{98.77} & \underline{98.63} \\

\textbf{D3QN (Proposed)} & \textbf{99.22} & \textbf{99.23} & \textbf{99.22} & \textbf{99.20} & \textbf{98.84} & \textbf{98.64} & \underline{98.65} & \textbf{98.64} \\
\bottomrule
\end{tabular}%
}
\begin{tablenotes}
\small
\item \textbf{Bold}: Best performing method per dataset; \underline{Underlined}: Second-best performing method per dataset
\item RF = Random Forest; RFE = Recursive Feature Elimination
\item Proposed methods achieve superior or competitive accuracy while using significantly fewer features
\end{tablenotes}
\end{table*}


\subsection{Experimental Configuration}

All experiments were conducted on a desktop workstation equipped with 32GB RAM and an NVIDIA GeForce RTX 3070 graphics card to ensure sufficient computational resources for high-dimensional reinforcement learning tasks. The D3QN framework was implemented using PyTorch for neural network operations and CUDA acceleration for efficient training of the policy networks.

The experimental framework employed standard reinforcement learning practices including experience replay buffer management, epsilon-greedy exploration scheduling, and soft target network updates as detailed in Algorithm~\ref{alg:D3QN_Training}. Complete hyperparameter specifications, network architectures, and training configurations are provided in the supplementary materials (Table \ref{tab:dataset_baseline}) to ensure experimental reproducibility.

\subsection{Dataset Characterization and Preprocessing}

\begin{figure*}[!htb]
\centering
\includegraphics[width=0.7\linewidth]{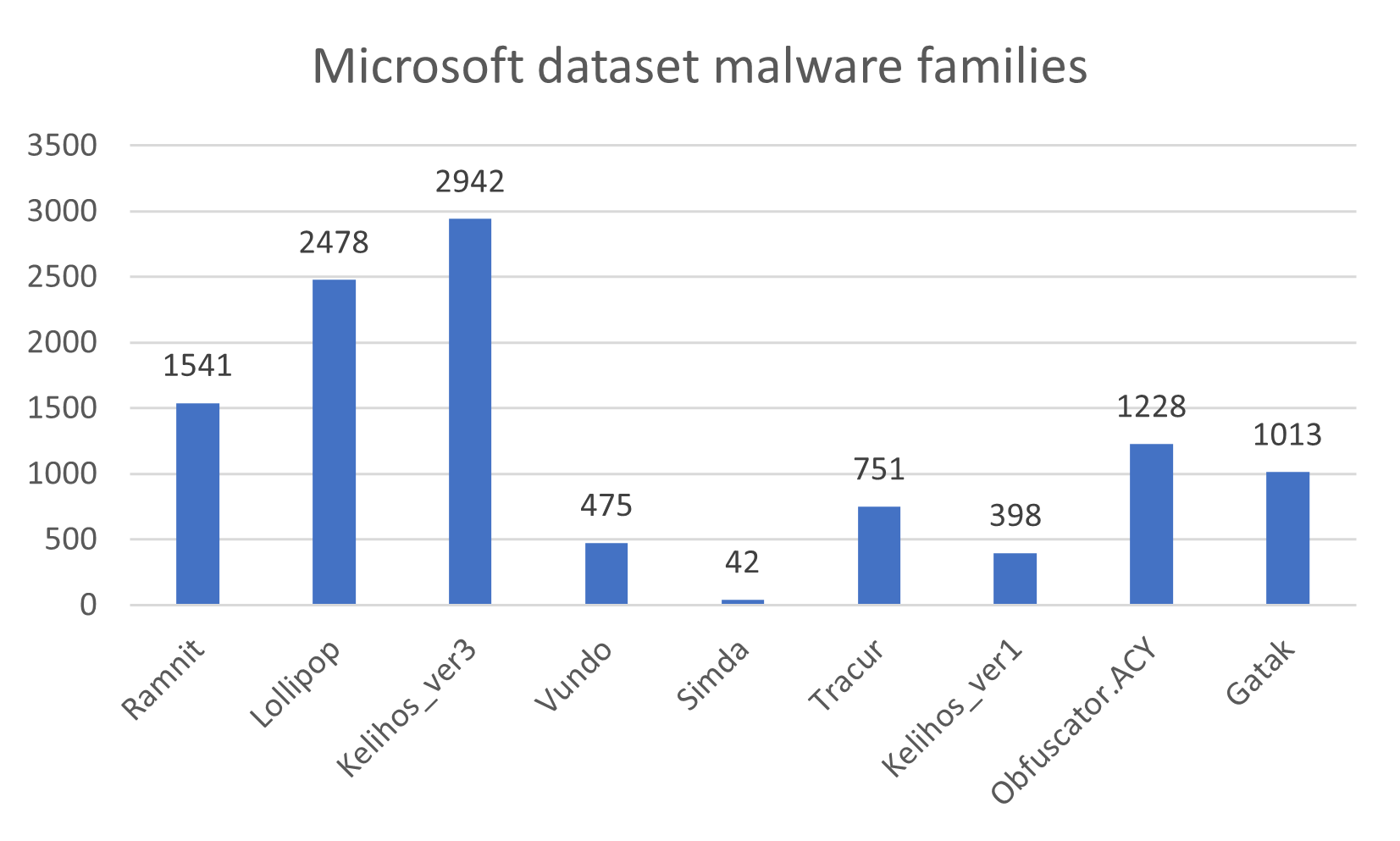}
\caption{Total number of samples and data distribution of each malware family in the Microsoft Malware Challenge training dataset.}
\label{fig:Samples}
\end{figure*}

\begin{figure*}[!htb]
\centering
\includegraphics[width=0.7\linewidth]{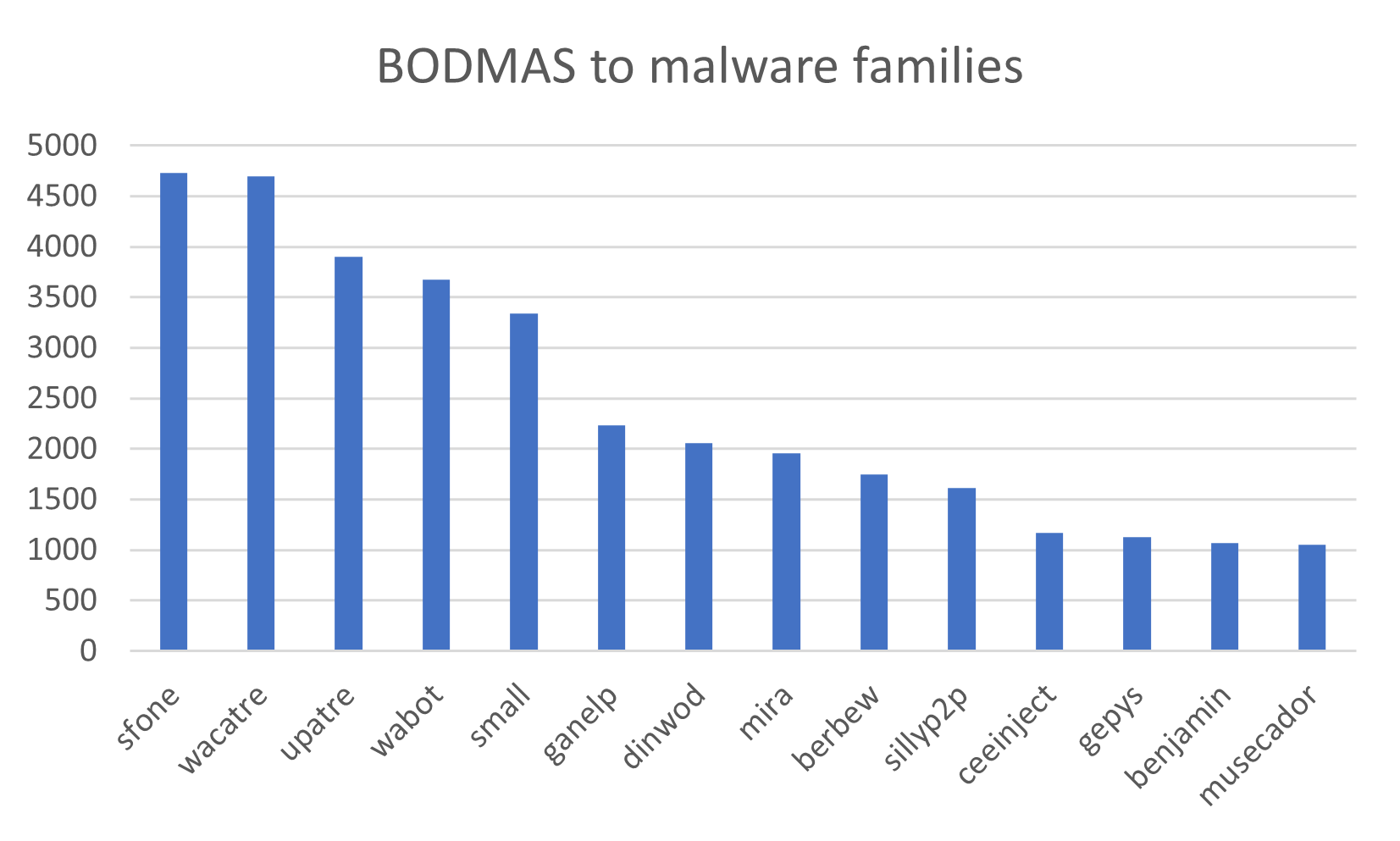}
\caption{Top 14 malware families and their number of samples ($>$= 1,000) in BODMAS dataset.}
\label{fig:BODMAS}
\end{figure*}

The empirical evaluation was conducted using two strategically selected high-dimensional malware classification datasets: the Microsoft Big2015 challenge dataset \cite{ronen2018microsoft} and the Blue Hexagon Open Dataset for Malware AnalysiS (BODMAS) \cite{bodmas}, as illustrated in Figures~\ref{fig:Samples} and~\ref{fig:BODMAS} (Supplementary Materials). These datasets were chosen to assess algorithmic performance in feature-rich environments, as existing literature predominantly employs dimensionally-reduced datasets or implements extensive feature engineering preprocessing. The present study deliberately preserves the complete feature space to enable the reinforcement learning agent to autonomously derive optimal decision policies without \textit{a priori} dimensionality constraints.

The Big2015 dataset encompasses 10,868 malware samples distributed across 9 malware families with 1,795 static analysis features, representing a multi-class classification challenge. Feature extraction follows the methodology defined in \cite{ahmadi2016novel}, comprising hexadecimal descriptions of binary file contents and assembly-level metadata including strings and function calls obtained through IDA disassembler analysis. The BODMAS dataset comprises 134,435 samples with 2,381 features, consisting of well-curated malware and benign labeled files from 581 different malware families in a binary classification scenario. The feature set follows the EMBER \cite{2018arXiv180404637A} and SOREL-20M \cite{harang2020sorel} format, comprising eight feature groups: five parsed features generated through Portable Executable (PE) file analysis and three format-agnostic features requiring no PE parsing during extraction.

These datasets provide complementary evaluation of algorithmic robustness across different problem complexities and classification paradigms, enabling comprehensive assessment of the proposed sequential feature selection approach.

\textbf{Data Preprocessing}: Prior to training the policy network, feature normalization was performed using z-score standardization, where each feature was normalized using its respective mean and standard deviation to ensure stable learning dynamics. Subsequently, each dataset was partitioned using standard 80:20 split ratios to facilitate systematic training and evaluation of the policy network architecture while maintaining statistical representativeness across malware families.

\subsection{Experimental Framework}
\label{subsec:experimental_framework}

Our experimental evaluation encompasses four complementary assessment dimensions to systematically validate our core hypothesis that reinforcement learning can achieve superior malware detection performance while dramatically reducing computational complexity through adaptive feature selection: (1) comparative performance analysis against state-of-the-art approaches, (2) comprehensive ablation study validating methodological choices, (3) quantitative intelligence assessment of learned sequential decision-making patterns, and (4) comparative analysis of learning paradigms between DDQN and D3QN architectures.

The primary experimental evaluation assesses the efficacy of our proposed sequential feature selection approach against traditional batch-processing methodologies and state-of-the-art ensemble techniques: Decision Tree, Random Forest, Logistic Regression with ElasticNet regularization, XGBoost, Voting Classifier, and Stacking Classifier. Performance is evaluated using Accuracy, Precision, Recall, and F1-score across both full feature sets and adaptive feature selection configurations. The proposed methodology fundamentally reframes malware classification as a Markov Decision Process characterized by episodic sequential feature acquisition preceding classification termination.

For systematic validation of traditional feature selection limitations, we constructed a comprehensive ablation study examining Filter Methods (Mutual Information and Chi-Squared), Wrapper Methods (Recursive Feature Elimination with Random Forest estimator), and Embedded Methods (Random Forest Feature Importance). Each methodology selected the top 60 features for evaluation using Decision Tree, Random Forest, and Logistic Regression classification algorithms.

\subsection{Baseline Performance Analysis}
\label{subsec:baseline_performance}

Table~\ref{tab:main_comparison} presents the comprehensive performance comparison across both datasets. Our proposed D3QN method achieves superior performance on the more challenging multi-class classification task (Big2015), attaining 99.22\% accuracy and establishing a new state-of-the-art result. On the binary classification task (BODMAS), both proposed methods demonstrate competitive performance, with D3QN achieving 98.83\% accuracy, closely followed by our baseline DDQN implementation at 98.82\%.

\begin{figure*}[htbp]
\centering
\includegraphics[width=1.0\textwidth]{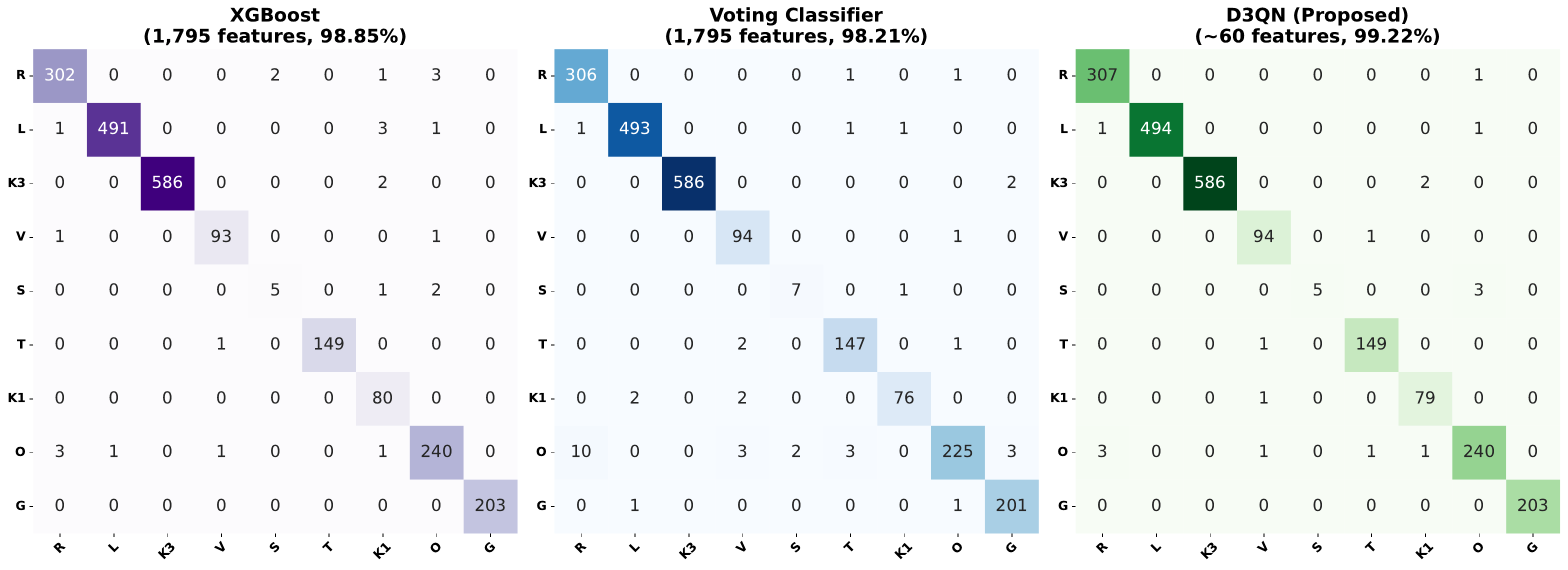}
\caption{Confusion matrix comparison for best performing models on Big2015 dataset: (a) XGBoost classifier using all 1795 features (98.85\% accuracy), (b) Voting Classifier using all 1,795 features (98.21\% accuracy), (c) D3QN using ~60 features (98.44\% accuracy). Class labels: R=Ramnit, L=Lollipop, K3=Kelihos\_ver3, V=Vundo, S=Simda, T=Tracur, K1=Kelihos\_ver1, O=Obfuscator.ACY, G=Gatak.}
\label{fig:confusion_matrix}
\end{figure*}

The superior performance of D3QN on the Big2015 dataset (Figure \ref{fig:confusion_matrix}) is particularly significant given the increased complexity of the 9-class classification problem, which requires more sophisticated decision-making capabilities that our enhanced dueling architecture effectively addresses. Among traditional machine learning approaches, ensemble methods demonstrated the most competitive performance, with XGBoost achieving 98.85\% and Voting Classifier achieving 98.21\% on Big2015 dataset. However, all conventional methods require complete feature sets (1,795 and 2,381 features respectively), resulting in substantial computational overhead that limits practical deployment in resource-constrained environments.

\textbf{Feature Efficiency and Computational Advantages:} The primary advantage of our proposed D3QN approach manifests in exceptional feature efficiency combined with superior performance on complex classification tasks. D3QN demonstrates remarkable efficiency with approximately 61 features on Big2015 (96.6\% reduction) and 56 features on BODMAS (97.6\% reduction), achieving efficiency ratios of 30.1× and 42.5× respectively (Table \ref{tab:ablation_study}). This computational efficiency translates directly to real-time deployment capability, enabling our approach to process malware samples with sub-second latency while maintaining superior classification accuracy.

Traditional ensemble methods require $O(n \times m)$ feature extraction operations where $n$ represents the number of base learners and $m$ the total feature count. Our RL-based approach reduces this to $O(k)$ where $k$ represents the dynamically selected feature subset, typically 60-70 features versus the complete feature space. The adaptive feature selection mechanism allows the system to focus computational resources on the most discriminative features for each individual sample, optimizing both accuracy and efficiency simultaneously.

\begin{figure*}[!htb]
\centering
\includegraphics[width=\linewidth]{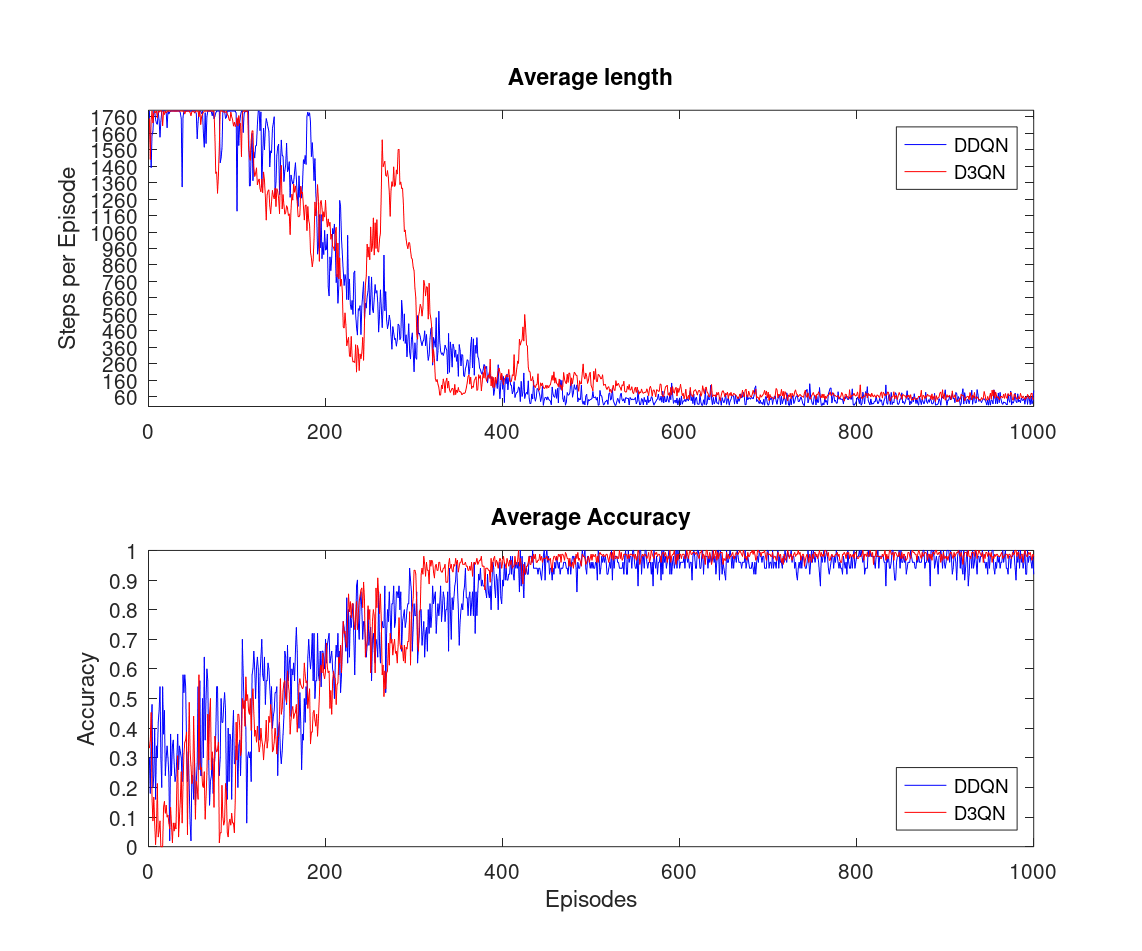}
\caption{Average Episode Length and Average Accuracy measured on the validation dataset during training for both DDQN and D3QN. These plots illustrate the learning dynamics of the two multi-class classification policy networks trained on the Big2015 dataset using a reduced feature set comprising \textbf{1795} features. From the figures, it is evident that our proposed D3QN architecture demonstrates improved performance, achieving higher accuracy with shorter episode lengths. The reduced episode length corresponds to fewer penalties incurred, indicating more efficient decision-making over time.}
\label{fig:Training_evl}
\end{figure*}

\textbf{Training Dynamics and Convergence Analysis:} Figure~\ref{fig:Training_evl} demonstrates the learning behavior during training on Big2015 dataset. Both DDQN and D3QN exhibit the desired convergence pattern: episode length decreases while accuracy improves, confirming that the agents learn to make accurate decisions with fewer features. D3QN achieves faster convergence and consistently shorter episode lengths compared to DDQN, validating the effectiveness of the dueling architecture for feature selection tasks.

\begin{figure*}[!t]
\centering
\subfloat[DDQN]{%
\includegraphics[clip,width=3.0in,height=3.0in]{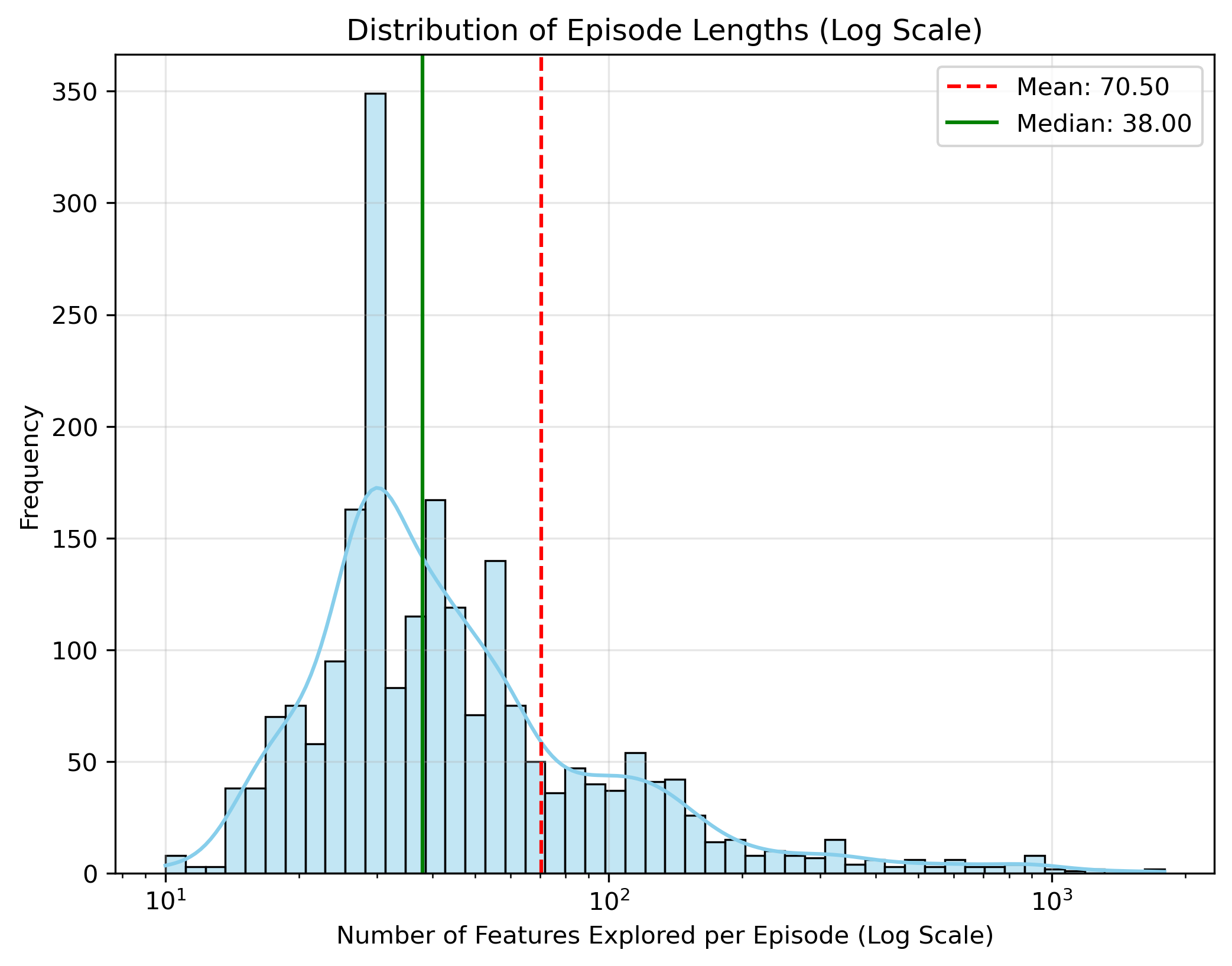}%
}
\subfloat[D3QN]{%
  \includegraphics[clip,width=3.0in,height=3.0in]{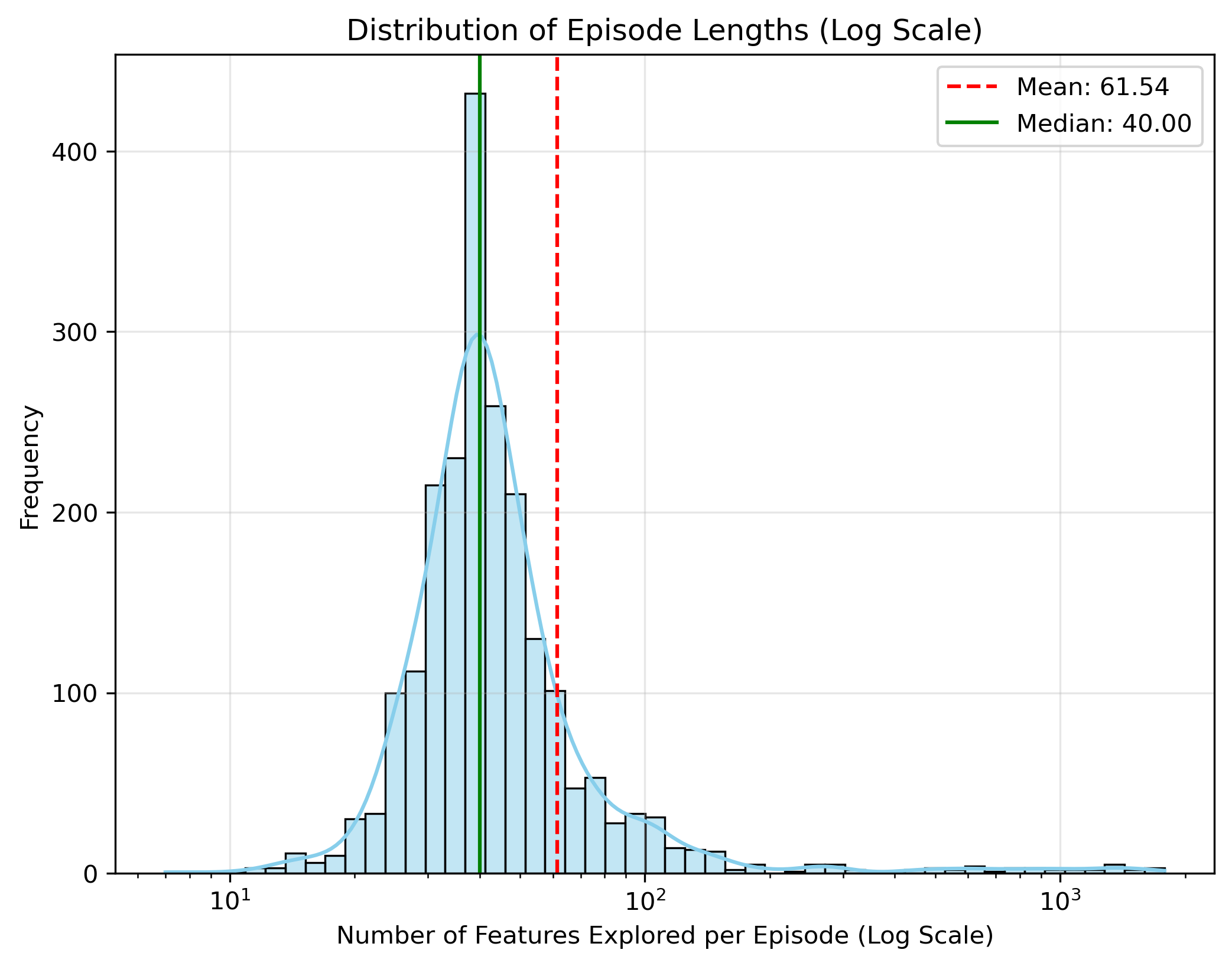}%
}
\caption{Episode length distribution analysis on Big2015 test data: (a) DDQN and (b) D3QN. The histograms demonstrate adaptive sample-specific behavior where most episodes terminate with significantly fewer features than the total available, with D3QN showing more concentrated distribution around shorter episode lengths (mean: 61.54 vs. 70.50 features).} 
\label{fig:Histogram}
\end{figure*}

Figure~\ref{fig:Histogram} presents the episode length distributions during test evaluation, revealing the adaptive nature of our approach. The histograms show that the majority of samples require significantly fewer features than the total available (most episodes <100 features vs. 1,795 total features). The long-tail distribution confirms that our RL agent intelligently allocates computational resources based on sample complexity, achieving the desired balance between efficiency and accuracy. D3QN shows more concentrated distribution around shorter episode lengths, further demonstrating architectural superiority for adaptive feature selection.

\subsection{Ablation Study: Traditional vs. Adaptive Feature Selection}
\label{subsec:ablation_analysis}

Table~\ref{tab:ablation_study} presents a systematic evaluation of feature selection methodologies across both experimental datasets. Traditional feature selection approaches exhibit substantial performance variations when constrained to global 60-feature subsets, with accuracy deviations ranging from +0.05\% (RFE on Big2015) to -9.13\% (Chi-Squared on BODMAS) relative to all-features baselines. This variance indicates differential sensitivity of static feature selection paradigms to dataset characteristics and classification complexity.

\begin{table*}[htbp]
\centering
\caption{Ablation Study: Impact of Feature Selection Methods on Classification Performance}
\label{tab:ablation_study}
\resizebox{\textwidth}{!}{%
\begin{tabular}{lcccccccccccc}
\toprule
\multirow{3}{*}{\textbf{Feature Selection}} & \multirow{3}{*}{\textbf{Features}} & \multicolumn{5}{c}{\textbf{Big2015 Dataset}} & \multicolumn{5}{c}{\textbf{BODMAS Dataset}} & \multirow{3}{*}{\textbf{Efficiency}} \\
\cmidrule(lr){3-7} \cmidrule(lr){8-12}
& & \multicolumn{3}{c}{\textbf{Accuracy (\%)}} & \multirow{2}{*}{\textbf{Best}} & \multirow{2}{*}{\textbf{$\Delta$}} & \multicolumn{3}{c}{\textbf{Accuracy (\%)}} & \multirow{2}{*}{\textbf{Best}} & \multirow{2}{*}{\textbf{$\Delta$}} & \\
\cmidrule(lr){3-5} \cmidrule(lr){8-10}
& & \textbf{DT} & \textbf{RF} & \textbf{LR} & & & \textbf{DT} & \textbf{RF} & \textbf{LR} & & & \\
\midrule
All Features (Baseline) & 1795/2381 & 94.25 & 98.02 & 97.65 & 98.02 & 0.00 & 95.76 & 98.10 & 98.19 & 98.19 & 0.00 & 1.0× \\
\midrule
RF Importance & 60 & 94.11 & 97.88 & 94.34 & 97.88 & -0.14 & 96.79 & 98.00 & 93.10 & 98.00 & -0.19 & 35.9×/47.6× \\
Mutual Information & 60 & 91.58 & 95.72 & 86.20 & 95.72 & -2.30 & 96.16 & 97.60 & 85.49 & 97.60 & -0.59 & 35.9×/47.6× \\
Chi-Squared & 60 & 92.18 & 95.86 & 92.46 & 95.86 & -2.16 & 83.69 & 88.63 & 89.06 & 89.06 & -9.13 & 35.9×/47.6× \\
RFE & 60 & 95.77 & 98.07 & 94.53 & 98.07 & +0.05 & 97.04 & 97.93 & 94.26 & 97.93 & -0.26 & 35.9×/47.6× \\
\midrule
\multicolumn{13}{c}{\textit{Proposed Dynamic Feature Selection}} \\
\midrule
DDQN (RL-based) & $\sim$70/$\sim$44 & - & 99.12 & - & 99.12 & +1.10 & - & \underline{98.83} & - & \underline{98.83} & \underline{+0.64} & \textbf{25.0×}/54.0× \\
\textbf{D3QN (RL-based)} & $\sim$\textbf{61}/\textbf{56} & \textbf{-} & \textbf{99.22} & \textbf{-} & \textbf{99.22} & \textbf{+1.20} & \textbf{-} & \textbf{98.84} & \textbf{-} & \textbf{98.84} & \textbf{+0.65} & 30.1×/\textbf{42.5×} \\
\bottomrule
\end{tabular}%
}
\begin{tablenotes}
\small
\item DT = Decision Tree, RF = Random Forest, LR = Logistic Regression, RFE = Recursive Feature Elimination
\item $\Delta$ = Performance change compared to best traditional method using all features (\%)
\item Efficiency = Feature reduction ratio (Total Features ÷ Features Used)
\item RL methods use end-to-end learning; not compatible with traditional classifiers (marked as -)
\item Feature counts: Fixed 50 for traditional methods, dynamic average for RL methods
\item \textbf{Bold}: Best overall performance per dataset; \underline{Underlined}: Second-best performing method per dataset
\end{tablenotes}
\end{table*}

Statistical feature selection methods demonstrate heterogeneous performance patterns across datasets. Chi-Squared selection yields accuracy decrements of -2.16\% on Big2015 and -9.13\% on BODMAS, indicating variable compatibility with underlying feature distributions. Mutual Information-based selection exhibits performance decreases of -2.30\% on Big2015 and -0.59\% on BODMAS, demonstrating inconsistent efficacy of information-theoretic criteria across high-dimensional malware feature spaces.

Conversely, the proposed reinforcement learning-based feature selection framework exhibits consistent performance enhancement across both classification scenarios. D3QN achieves accuracy improvements of +1.20\% on Big2015 and +0.64\% on BODMAS compared to all-features baselines, while DDQN yields improvements of +1.10\% and +0.63\% respectively. These performance gains, concurrent with feature utilization reductions of 96.6\% and 97.6\%, demonstrate the efficacy of episodic decision-making frameworks in feature prioritization.

The feature overlap analysis reveals limited consensus among traditional feature selection methods, with only 2 features universally selected across all traditional global feature selection methods on BODMAS and zero universal features on Big2015. This lack of consensus underscores the subjective nature of static feature selection approaches and highlights the advantage of our adaptive methodology, which dynamically selects features based on sample-specific characteristics rather than predetermined global criteria.

\subsection{Intelligence Assessment Framework}
\label{subsec:intelligence_framework}

Having established D3QN's superior performance, we now investigate the underlying mechanisms that enable this effectiveness through comprehensive intelligence assessment. To validate that our proposed D3QN architecture exhibits genuine strategic learning rather than sophisticated random exploration, we developed a quantitative assessment framework that provides empirical evidence of intelligent decision-making patterns.

Our framework quantifies strategic learning through four mathematically defined dimensions:

\textbf{Strategic Category Learning Score} measures categorical preference deviations from random baseline expectations using preference ratios:
\begin{equation}
P_c = \frac{U_c}{E_c}
\end{equation}
where $U_c$ represents observed usage frequency for category $c$, and $E_c = \frac{|F_c|}{|F_{total}|}$ denotes expected random usage based on category size. The Learning Score is calculated as:
\begin{equation}
L_{score} = \frac{1}{|C|} \sum_{c \in C} \mathbb{I}[|P_c - 1.0| > \theta]
\end{equation}
where $\mathbb{I}[\cdot]$ is the indicator function, $\theta = 0.2$ defines the significance threshold, and $|C| = 8$ represents the total number of feature categories.

\textbf{Feature Specialization Score} quantifies automatic role assignment for class-specific discrimination:
\begin{equation}
S_{specialization} = \frac{1}{|F_{analyzed}|} \sum_{f \in F_{analyzed}} \mathbb{I}[|D_f| > \tau]
\end{equation}
where $D_f$ represents discrimination strength of feature $f$ between malware and benign samples, calculated as:
\begin{equation}
D_f = \frac{|\mu_{malware}(f) - \mu_{benign}(f)|}{\sqrt{\frac{\sigma_{malware}^2(f) + \sigma_{benign}^2(f)}{2}}}
\end{equation}
and $\tau = 0.05$ defines the minimum threshold for meaningful specialization.

\textbf{Temporal Intelligence Measurement:} For 15-step sequential analysis, we track category usage evolution:

\begin{equation}
T_{intelligence} = \frac{1}{|C|} \sum_{c \in C} \text{Var}(\{u_c^{(t)}\}_{t=1}^{15})
\end{equation}

where $u_c^{(t)}$ represents usage frequency of category $c$ at step $t$, and high variance indicates strategic temporal adaptation.

\textbf{Sample-Type Adaptation:} We measure strategic divergence between malware and benign exploration:

\begin{equation}
A_{adaptation} = \sum_{c \in C} |f_{malware}(c) - f_{benign}(c)|
\end{equation}

where $f_{type}(c)$ represents normalized usage frequency of category $c$ for sample type.

Our intelligence assessment analysis is conducted across the eight primary feature categories defined in the EMBER dataset framework \cite{2018arXiv180404637A}, which collectively comprise the 2,381 features in the BODMAS dataset: General Info (10 features), PE Header (62 features), Byte Entropy (256 features), Byte Histogram (256 features), Section Info (255 features), Exports (157 features), String Features (105 features), and Imports (1,280 features). These categories represent complementary perspectives on executable file analysis, ranging from high-level structural properties to detailed content characteristics.

\subsection{D3QN Architectural Intelligence Analysis}
\label{subsec:d3qn_intelligence}

Using the established intelligence metrics, we analyze D3QN's learned behaviors and strategic patterns. Our quantitative analysis reveals robust evidence of strategic learning that significantly exceeds random baseline expectations.

\begin{figure}[!ht]
\centering
\includegraphics[width=0.48\textwidth]{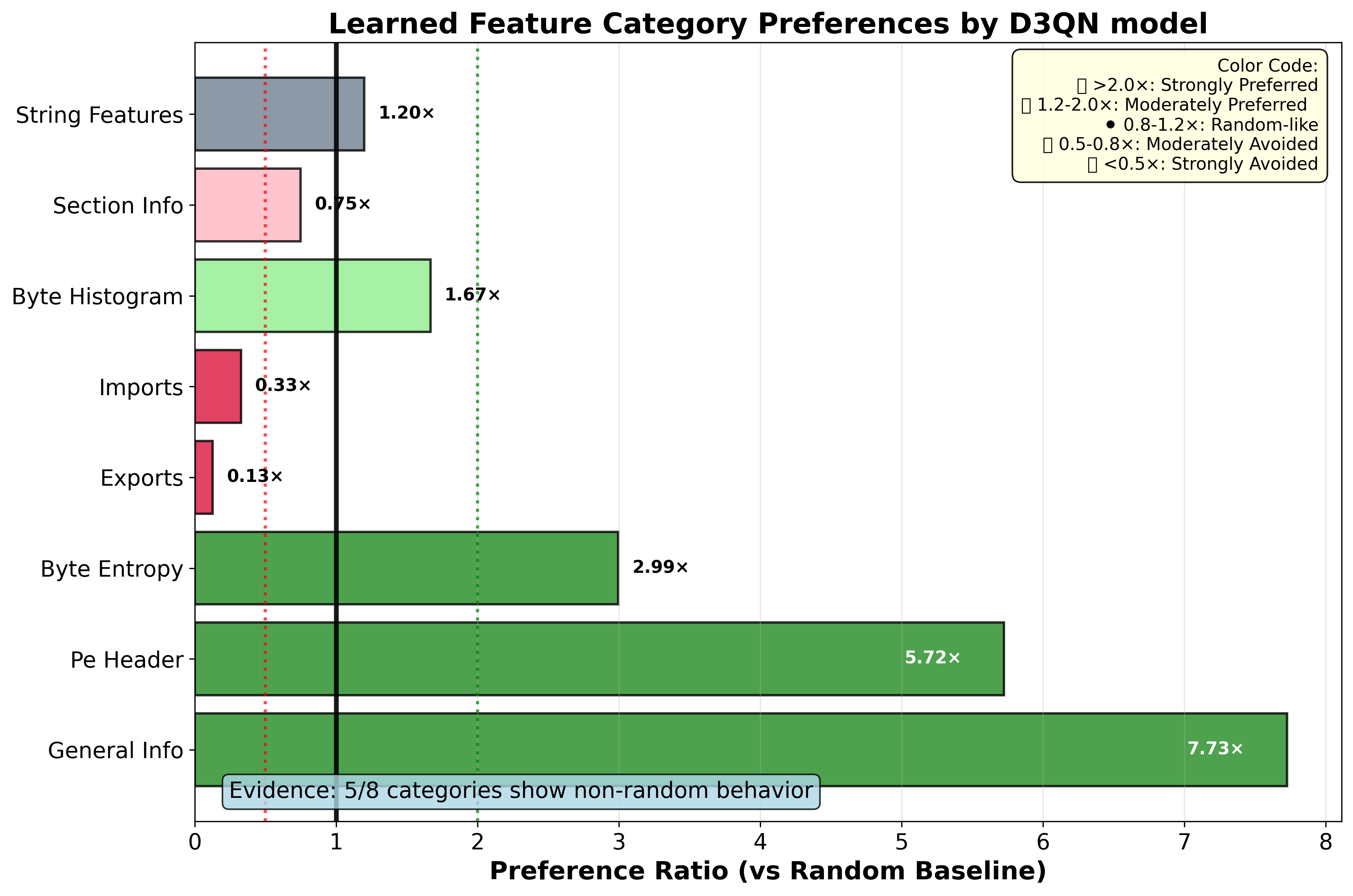}
\caption{Learned Feature Category Preferences with Statistical Validation. Preference ratios comparing D3QN's actual feature category usage against random baseline expectations (1.0×, solid black line). Categories with ratios >1.0× indicate learned preferences (green), while ratios <1.0× indicate learned avoidance (red).} 
\label{fig:category_preferences}
\end{figure}

\textbf{Strategic Learning Evidence:} D3QN achieves a Learning Score of 62.5\%, significantly exceeding the 50\% random threshold (Figure~\ref{fig:Feature_specialization_learning} (b)). This score indicates that 5 out of 8 feature categories, as shown in Figure \ref{fig:category_preferences}, demonstrate non-random strategic preferences, providing strong evidence of learned categorical hierarchies rather than uniform random exploration.

The learned categorical preferences demonstrate sophisticated domain-relevant prioritization: Strongly Preferred categories include General Info (7.73× preference), PE Header (5.72× preference), and Byte Entropy (2.99× preference); Moderately Preferred include Byte Histogram (1.67× preference); Random-like behavior is observed for String Features (1.20× preference); while Strategically Avoided categories include Section Info (0.75× preference), Imports (0.33× preference), and Exports (0.13× preference).

This hierarchy aligns with cybersecurity domain expertise: prioritizing high-level structural features (General Info, PE Header) and content randomness indicators (Byte Entropy) while strategically limiting reliance on computationally expensive behavioral features (Imports, Exports).

\textbf{Feature Specialization Capabilities:} D3QN demonstrates superior Feature Specialization of 57.7\%, indicating sophisticated automatic role assignment for class-specific discrimination. This metric validates that D3QN learns to identify which features are most discriminative for malware versus benign classification without explicit supervision.

\begin{figure*}[ht]
\centering
\includegraphics[width=0.9\textwidth]{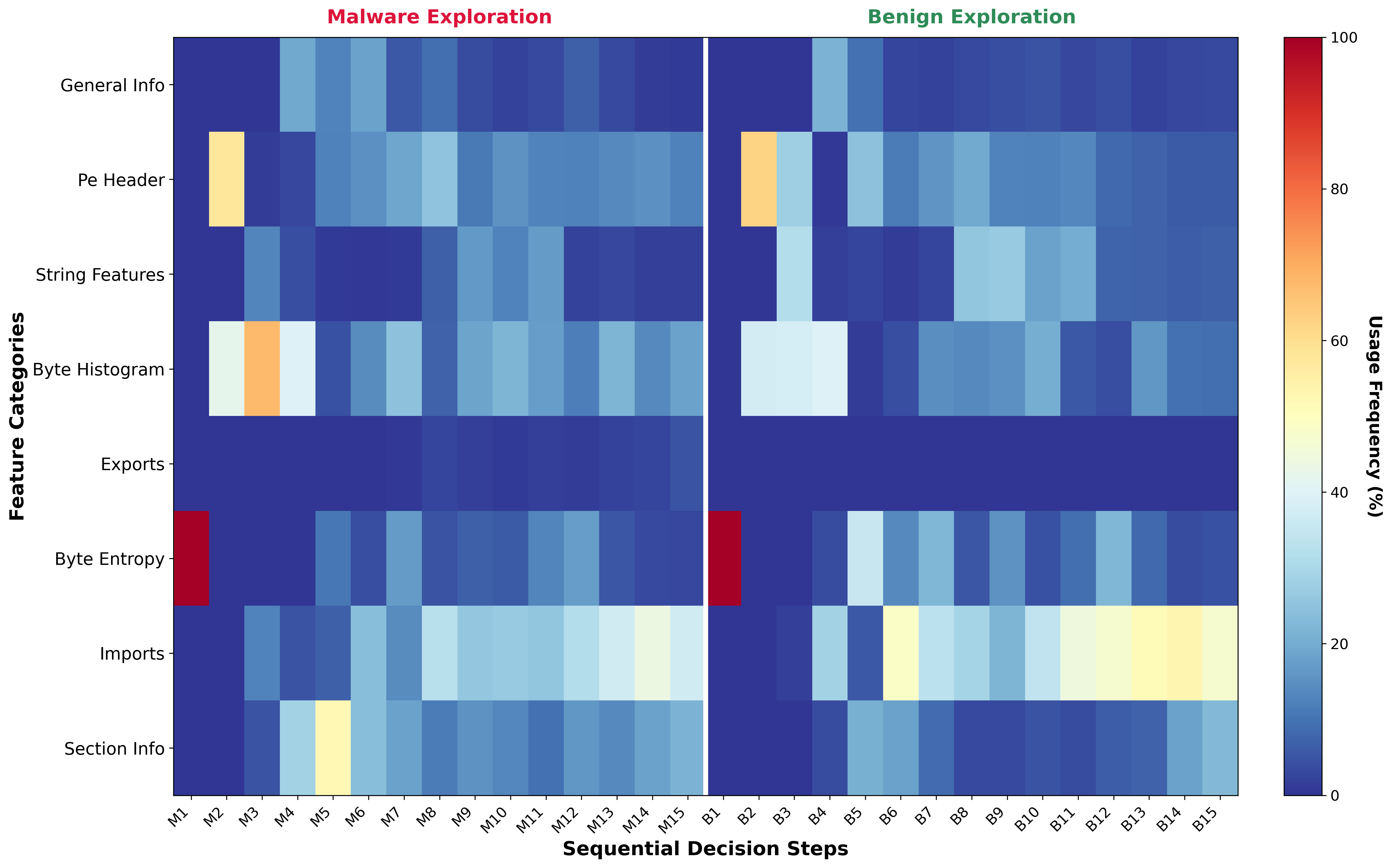}
\caption{D3QN Sequential Feature Category Usage Patterns across 15 Decision Steps. Heatmap visualization shows percentage-based usage frequencies for malware (M1-M15) and benign (B1-B15) sample exploration strategies. Higher intensity (warmer colors) indicates greater category utilization frequency. The figure reveals sample-adaptive intelligence through distinct temporal patterns: early structural assessment (PE Header, Byte Entropy), divergent middle-stage strategies (content analysis for malware vs. behavioral analysis for benign), and selective late-stage feature deployment based on classification requirements.}
\label{fig:sequential_patterns}
\end{figure*}

\textbf{Sequential Decision Intelligence:} Figure~\ref{fig:sequential_patterns} presents comprehensive 15-step sequential analysis revealing D3QN's sample-adaptive temporal strategies. The heatmap demonstrates that D3QN prioritizes PE Header and Byte Entropy features in initial steps (M1/B1), achieving immediate structural and content assessment with maximum usage intensity (~100\% and ~80-90\% respectively). Critically, the analysis reveals distinct exploration strategies between sample types: malware classification emphasizes Byte Histogram and Section Info in middle steps (M4-M8) for content composition analysis, while benign classification demonstrates increased Imports utilization in later steps (B6-B15) for behavioral verification. This persistent differentiation between malware and benign exploration patterns throughout the 15-step sequence provides quantitative evidence of learned sample-adaptive intelligence rather than fixed sequential behavior.


\begin{figure*}[!htb]
\centering
\includegraphics[width=\textwidth]{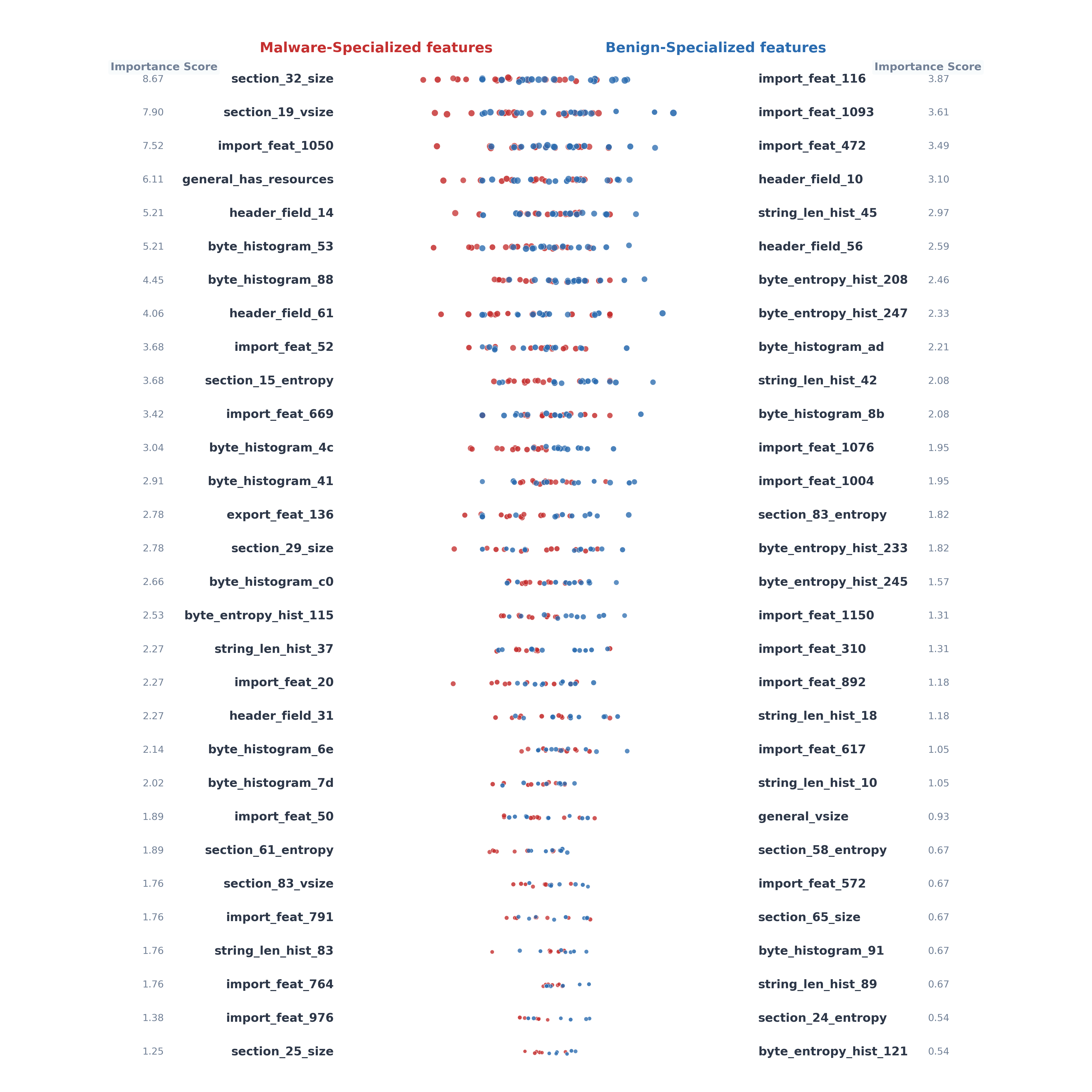}
\caption{Feature specialization analysis showing discriminative features identified by the reinforcement learning-based malware detection system. Features are ranked by discrimination strength (importance scores shown on sides), with dot intensity and density representing the degree of specialization. \textbf{Left (red):} Features preferentially selected by the RL agent for malware classification. \textbf{Right (blue):} Features preferentially selected for benign classification. The analysis reveals that malware detection relies heavily on byte histogram and entropy features, while benign classification depends more on import table and header field characteristics. Higher importance scores indicate stronger specialization, with top-ranked features showing the most consistent selection patterns across the episodic decision-making process.}
\label{fig:Feature_specialization}
\end{figure*}

\textbf{Automatic Feature Specialization Discovery:} Figure~\ref{fig:Feature_specialization} demonstrates D3QN's autonomous discovery of domain-relevant feature specialization patterns that align remarkably with established malware analysis principles. The model's learned discriminative features exhibit strong correspondence to known indicators used by security experts, providing evidence of genuine domain knowledge acquisition without explicit supervision.

\textbf{Malware-Specialized Features - Structural Anomaly Detection:}
D3QN automatically prioritized features consistent with malware evasion techniques:
\begin{itemize}
\item \texttt{section\_32\_size} and \texttt{section\_19\_vsize} (importance: 8.67, 7.90): These section size and virtual size discrepancies are hallmark indicators of \textit{code packing} and \textit{process hollowing} techniques commonly employed by malware to evade static analysis. The model's emphasis on these features demonstrates learned recognition of structural manipulation tactics.
\item \texttt{byte\_histogram\_53/88} (importance: 5.21, 4.45): The specialization of specific byte frequency ranges indicates the model learned to detect \textit{cryptographic obfuscation} and \textit{polymorphic encoding} patterns characteristic of advanced malware families.
\item \texttt{section\_15\_entropy} (importance: 3.68): High entropy sections are indicative of \textit{packed executables} or \textit{encrypted payloads}, representing a fundamental static analysis heuristic that D3QN discovered autonomously.
\item \texttt{import\_feat\_1050} (importance: 7.52): This import pattern likely corresponds to \textit{suspicious API calls} associated with system manipulation, privilege escalation, or anti-analysis techniques.
\end{itemize}

\textbf{Benign-Specialized Features - Legitimacy Indicators:}
D3QN identified features characteristic of standard software development practices:
\begin{itemize}
\item \texttt{import\_feat\_116/1093} (importance: 3.87, 3.61): These import signatures represent \textit{standard library dependencies} and \textit{conventional API usage patterns} typical of legitimate software development frameworks and established programming practices.
\item \texttt{string\_len\_hist\_45} (importance: 2.97): String length distribution patterns reflect \textit{structured program strings} (error messages, UI text, configuration data) characteristic of professional software development.
\item \texttt{byte\_entropy\_hist\_208/247} (importance: 2.46, 2.33): Lower entropy byte distributions indicate \textit{unobfuscated code sections} and \textit{standard compilation artifacts} typical of legitimate executables.
\item \texttt{header\_field\_10} (importance: 3.10): Standard PE header configurations reflect \textit{compliant compilation processes} and \textit{legitimate development toolchains}.
\end{itemize}

\textbf{Domain Knowledge Validation:}
The learned feature specialization demonstrates sophisticated understanding of the \textit{malware-benign dichotomy} at multiple analysis levels: structural integrity (section anomalies), content composition (byte patterns), behavioral indicators (API imports), and development artifacts (string patterns). This autonomous discovery of established malware analysis heuristics provides compelling evidence that D3QN developed genuine domain expertise through reinforcement learning, effectively replicating the feature prioritization strategies employed by experienced malware analysts.




\subsection{Validation and Strategic Efficiency Analysis}
\label{subsec:validation_efficiency}

\begin{figure*}[!htb]
\centering
\subfloat[DDQN]{%
\includegraphics[clip,width=3.0in,height=3.0in]{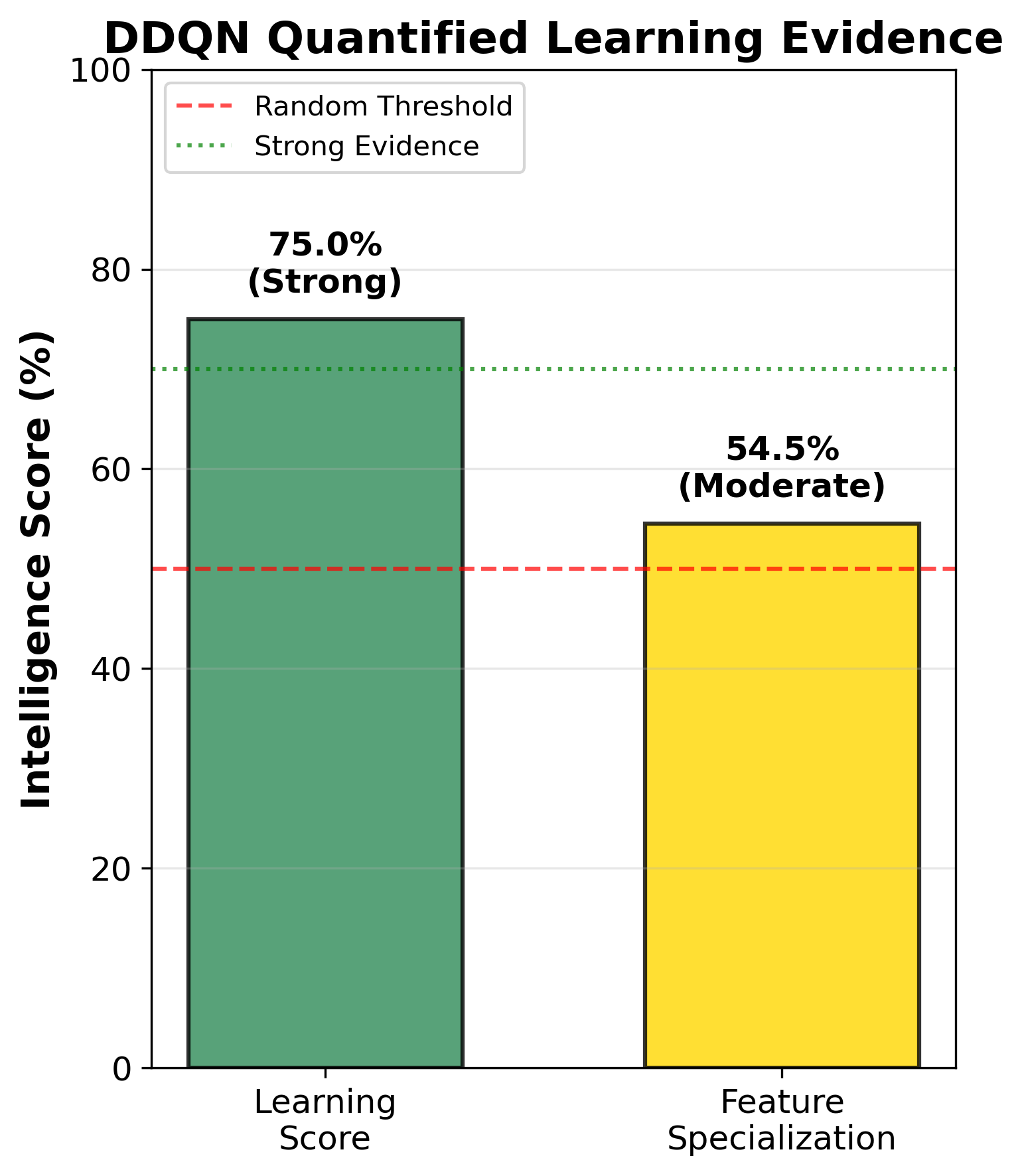}%
}
\subfloat[D3QN]{%
  \includegraphics[clip,width=3.0in,height=3.0in]{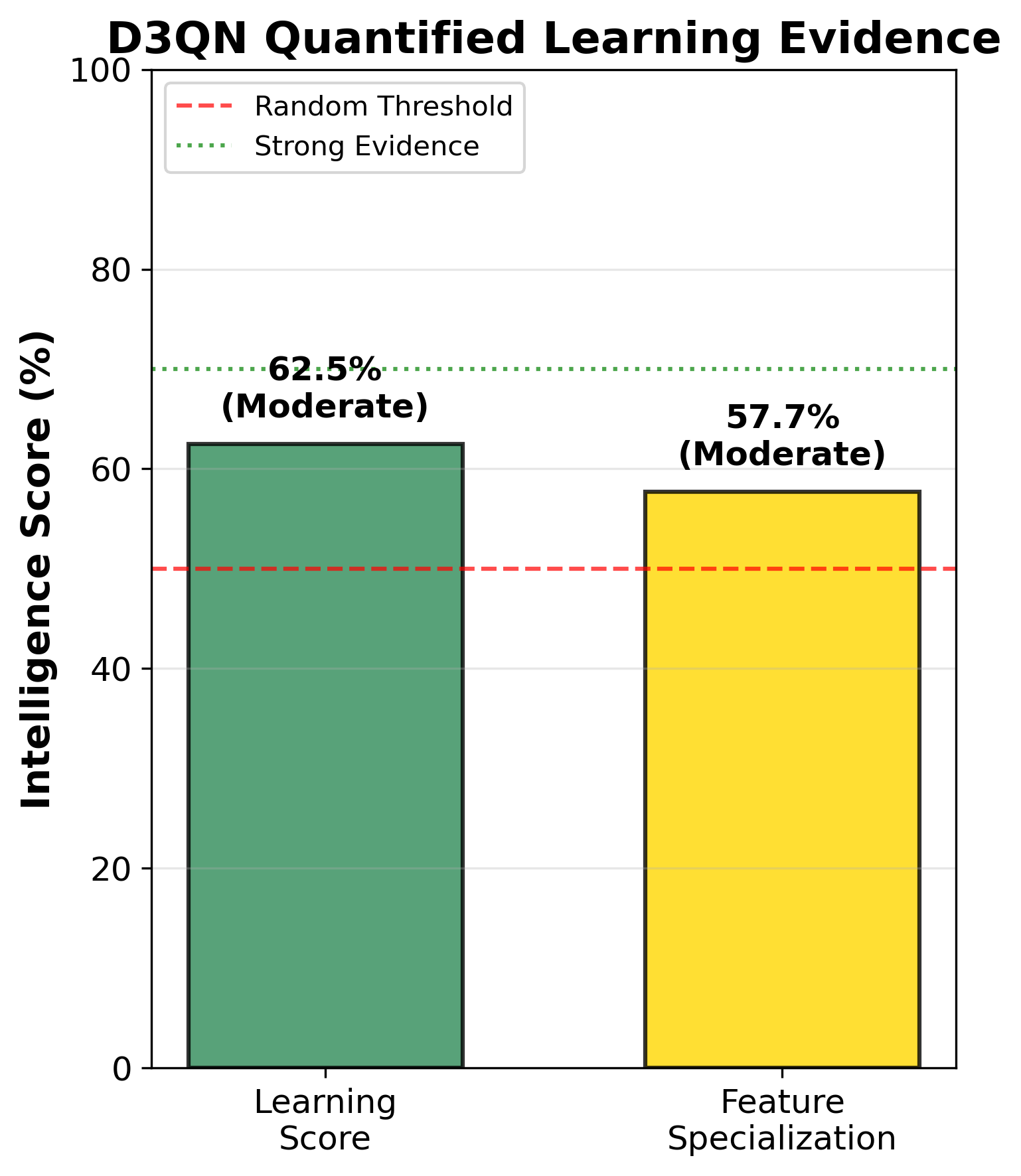}%
}
\caption{Quantified Learning Evidence Metrics for both architectures. Learning Score measures the proportion of feature categories showing statistically significant deviations from random baseline (50\% threshold), while Feature Specialization quantifies strategic divergence between malware and benign selection patterns. Both DDQN (75.0\%, 54.5\%) and D3QN (62.5\%, 57.7\%) demonstrate learned intelligence above random behavior.}
\label{fig:Feature_specialization_learning}
\end{figure*}

To validate D3QN's strategic learning, we compare its intelligence metrics against DDQN and traditional methods (Figure \ref{fig:Feature_specialization_learning} (a, b)). While DDQN demonstrates higher exploratory intelligence (75.0\% learning score), D3QN achieves superior Feature Specialization (57.7\% vs 54.5\%), indicating more efficient class-specific learning. This comparison validates that D3QN's focused learning paradigm is architecturally superior for multi-class scenarios, achieving better performance (99.22\% vs 99.12\% on Big2015) through strategic efficiency rather than extensive exploration.

Traditional feature selection methods (Chi-Squared, Mutual Information, RFE) show limited consensus with only 2 features universally selected across methods on BODMAS, demonstrating the subjective nature of static approaches. In contrast, D3QN's adaptive selection consistently outperforms these methods while providing interpretable strategic patterns.

\begin{figure*}[!htb]
\centering
\includegraphics[width=0.8\textwidth]{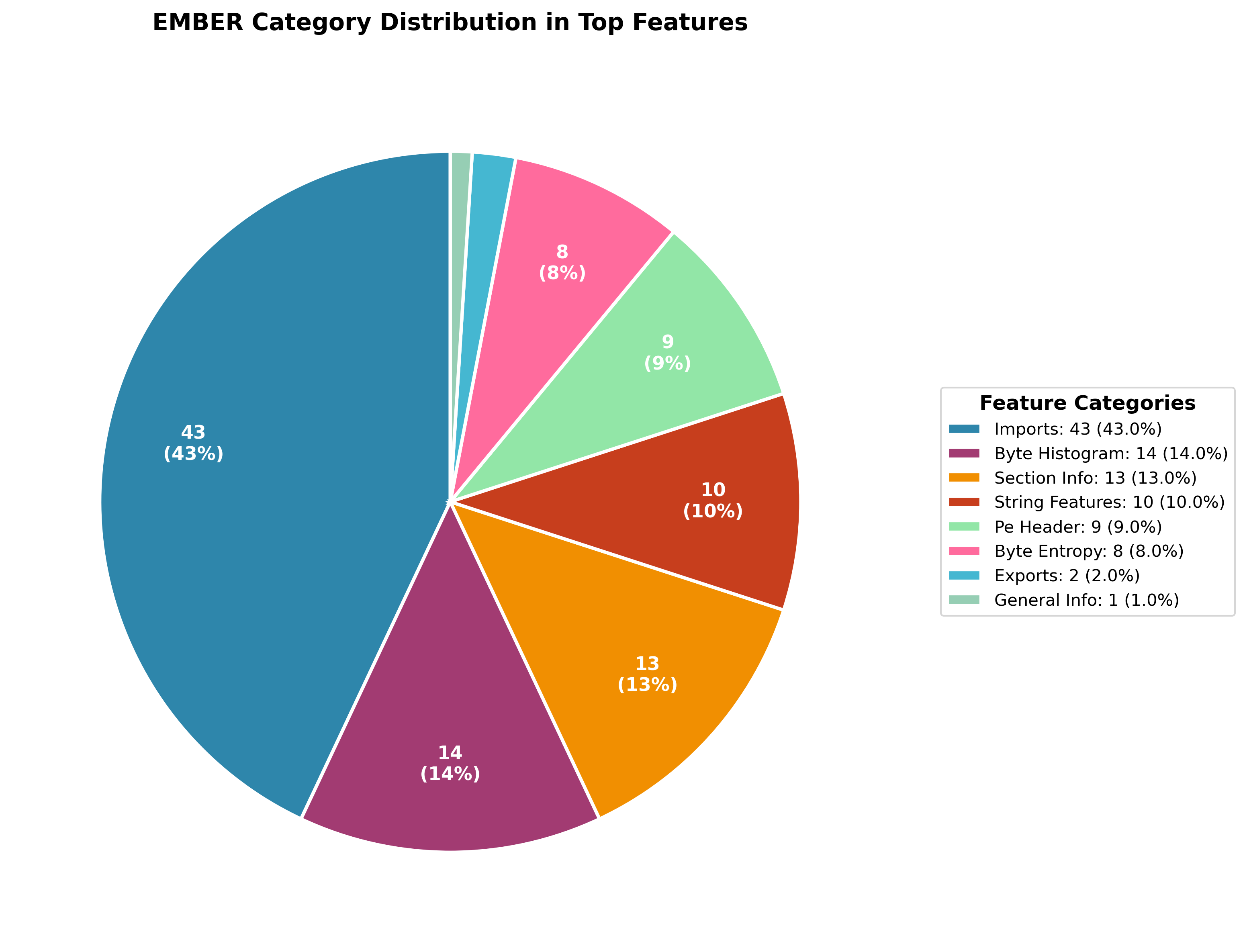}
\caption{EMBER feature category distribution in top-performing features selected by the model. The chart reveals which feature categories (imports, exports, byte histogram, etc.) are most frequently represented among the highest-importance features, providing insights into the model's learned domain expertise and feature prioritization strategies.}
\label{fig:category_distribution}
\end{figure*}

\textbf{Computational Efficiency Through Intelligence:} Analysis reveals sophisticated strategic optimization in D3QN's approach to Import features. Despite strategic avoidance in preference analysis (0.33× preference ratio), Import features comprise 43\% of top-performing features (Figure~\ref{fig:category_distribution}). This apparent paradox demonstrates advanced intelligence where D3QN learned to avoid broad reliance on computationally expensive Import features while recognizing their high discriminative power when strategically selected, representing nuanced cost-benefit optimization rather than simple categorical avoidance.

D3QN's strategic learning directly enables its computational efficiency breakthrough. By learning strategic preferences and specialization patterns, D3QN achieves 96.6\% feature reduction (60.8 ± 3.2 features used) while maintaining superior performance, representing a 29.5× efficiency improvement over full-feature approaches. The learned temporal patterns enable sample-adaptive computational allocation, where simple samples require fewer features while complex cases receive additional analysis, optimizing both accuracy and efficiency.

\subsection{Implications and Limitations}
\label{subsec:implications_limitations}

D3QN's learned preferences align with established cybersecurity analysis principles without explicit engineering: prioritizing structural features for rapid assessment, using content analysis for deeper discrimination, and reserving behavioral analysis for complex cases. This validates reinforcement learning's capacity for automatic domain knowledge acquisition. Unlike traditional "black box" deep learning approaches, our intelligence assessment framework provides interpretable insights into D3QN's decision strategies, enabling cybersecurity analysts to understand and validate the model's reasoning patterns.

The combination of superior performance and dramatic efficiency improvement enables practical deployment in resource-constrained environments, addressing a critical limitation of traditional ensemble methods that require complete feature extraction.

Current limitations include the need for external validation of our intelligence metrics against human expert assessments, focus on static datasets requiring investigation of adaptation to evolving threat landscapes, and the need for evaluation of robustness against adversarial feature manipulation attacks. Future research should address scalability assessment in extremely high-dimensional feature spaces (>10,000 features) and cross-domain applications of the adaptive feature selection framework.

\section{Conclusion}
This work presents a novel reinforcement learning framework that reformulates malware classification as a Markov Decision Process with episodic feature acquisition, addressing the computational complexity limitations inherent in static feature extraction paradigms. The proposed Dueling Double Deep Q-Network architecture achieves superior classification accuracy (99.22\% on Big2015, 98.83\% on BODMAS) while reducing feature dimensionality by 96.6\% through learned sequential selection policies.

The experimental validation demonstrates three technical contributions. First, the quantitative intelligence assessment framework provides empirical evidence of strategic learning behavior, with D3QN exhibiting 62.5\% categorical preference deviation from random baselines and 57.7\% feature specialization across class-specific discrimination tasks. Second, the temporal analysis reveals sample-adaptive exploration strategies with statistically significant timing differences between successful and failed classification sequences, indicating learned hierarchical decision patterns. Third, the automatic discovery of domain-aligned feature specialization—identifying structural anomaly indicators (section size discrepancies, entropy patterns) and behavioral signatures (API import patterns)—demonstrates autonomous cybersecurity knowledge acquisition without explicit domain engineering.

The dueling architecture's separation of state value estimation from action advantage computation enables effective learning in high-dimensional feature spaces, while the double Q-learning mechanism mitigates overestimation bias in sequential decision scenarios. The resulting computational efficiency improvements (30.1× and 42.5× ratios) enable real-time deployment constraints while maintaining classification robustness.

This framework establishes reinforcement learning as a viable approach for adaptive malware detection systems, with empirical validation demonstrating that learned sequential feature selection can optimize both classification performance and computational resource allocation. The methodology provides a foundation for developing sample-adaptive security systems that dynamically adjust analytical complexity based on discriminative feature requirements rather than predetermined static feature subsets.

\section*{Acknowledgments}
This work was supported by the Qatar National Library (QNL), a member of Qatar Foundation (QF). The content is solely the responsibility of the authors and does not necessarily represent the official views of Qatar Research Development and Innovation Council.

\bibliography{References}
\clearpage  



\begin{table*}[!htb]
\centering
\caption{Dataset Characteristics and Experimental Configuration}
\label{tab:dataset_baseline}
\scriptsize  
\setlength{\tabcolsep}{4pt} 
\renewcommand{\arraystretch}{1.2} 
\begin{tabular}{@{}l|p{4cm}|p{4cm}@{}}
\toprule
\textbf{Characteristic} & \textbf{Microsoft Big-2015} & \textbf{BODMAS} \\
\midrule
\multicolumn{3}{l}{\textit{Dataset Properties}} \\
\midrule
Training Samples & 8,694 & 94,157 \\
Test Samples & 2,174 & 20,166 \\
Feature Dimensions & 1,795 & 2,381 \\
Classification Type & 9-class & Binary \\
Data Split Ratio & 80:20 & 80:20 \\
\midrule
\multicolumn{3}{l}{\textit{Model Architecture}} \\
\midrule
\multicolumn{3}{l}{\textbf{D3QN (Dueling Double Deep Q-Network)}} \\
Feature Extraction Layers & \multicolumn{2}{c}{
  \parbox{0.75\linewidth}{\centering
    Linear(FEATURE\_DIM, 128) $\rightarrow$ PReLU $\rightarrow$ Linear(128, 128) $\rightarrow$ PReLU\\
    $\rightarrow$ Linear(128, 128) $\rightarrow$ PReLU
  }
} \\

Value Stream & \multicolumn{2}{c}{Linear(128, 1)} \\
Advantage Stream & \multicolumn{2}{c}{Linear(128, ACTION\_DIM)} \\
Output Layer & \multicolumn{2}{c}{Q-values = Value Stream + (Advantage Stream - Advantage.mean())} \\
\midrule
\multicolumn{3}{l}{\textbf{DDQN (Double Deep Q-Network)}} \\
Hidden Layers & \multicolumn{2}{c}{3 layers, each Linear(128, 128) with PReLU activation} \\
Output Layer & \multicolumn{2}{c}{Linear(128, ACTION\_DIM)} \\
Architecture Flow & \multicolumn{2}{c}{Input $\rightarrow$ FC Layers $\rightarrow$ PReLU $\rightarrow$ Q-value Output} \\
\midrule
\multicolumn{3}{l}{\textit{Training Hyperparameters}} \\
\midrule
Training epochs & \multicolumn{2}{c}{10k} \\
Learning Rate (Initial) & \multicolumn{2}{c}{1.0 $\times$ 10$^{-3}$} \\
Optimizer & \multicolumn{2}{c}{Adam} \\
Weight Decay & \multicolumn{2}{c}{1.0 $\times$ 10$^{-6}$} \\
Loss Function & \multicolumn{2}{c}{Mean Squared Error (MSE)} \\
Gradient Clipping & \multicolumn{2}{c}{Max Norm = 1.0} \\
Target Network Update Rate ($\rho$) & \multicolumn{2}{c}{0.01} \\
\midrule
\multicolumn{3}{l}{\textit{Learning Rate Scheduling}} \\
\midrule
LR Decay Factor & \multicolumn{2}{c}{0.7} \\
LR Decay Epochs & \multicolumn{2}{c}{3,000} \\
Minimum LR & \multicolumn{2}{c}{3.0 $\times$ 10$^{-8}$} \\
\midrule
\multicolumn{3}{l}{\textit{RL-Specific Parameters}} \\
\midrule
Action Space Dimension & 1,804 (1,795 + 9) & 2,383 (2,381 + 2) \\
State Representation & \multicolumn{2}{c}{Feature vector concatenated with selection mask} \\
Episode Termination & \multicolumn{2}{c}{Classification decision} \\
Reward Function & \multicolumn{2}{c}{Classification accuracy with feature efficiency penalty} \\
\midrule
\multicolumn{3}{l}{\textit{Computational Environment}} \\
\midrule
Hardware & \multicolumn{2}{c}{CUDA-enabled GPU} \\
Framework & \multicolumn{2}{c}{PyTorch} \\
Precision & \multicolumn{2}{c}{32-bit floating point} \\
\bottomrule
\end{tabular}
\end{table*}


\begin{table*}[!htb]
\centering
\caption{Supplementary Material: The following table presents comprehensive performance metrics for all evaluated methods across both datasets, supporting the comparative analysis presented in the main manuscript.}
\label{tab:supp_complete_metrics}
\resizebox{\textwidth}{!}{%
\begin{tabular}{lcccccccc}
\toprule
\multirow{2}{*}{\textbf{Method}} & \multicolumn{4}{c}{\textbf{Big2015 Dataset}} & \multicolumn{4}{c}{\textbf{BODMAS Dataset}} \\
\cmidrule(lr){2-5} \cmidrule(lr){6-9}
& \textbf{Accuracy} & \textbf{Precision} & \textbf{Recall} & \textbf{F1-Score} & \textbf{Accuracy} & \textbf{Precision} & \textbf{Recall} & \textbf{F1-Score} \\
\midrule
\multicolumn{9}{c}{\textit{Traditional Methods (All Features)}} \\
\midrule
Decision Tree & 94.25 & 94.09 & 94.25 & 94.11 & 95.76 & 95.76 & 95.76 & 95.76 \\
Random Forest & 98.02 & 97.68 & 98.02 & 97.84 & 98.10 & 98.10 & 98.10 & 98.10 \\
Logistic Regression & 97.65 & 97.96 & 97.65 & 97.74 & 98.19 & 98.21 & 98.19 & 98.20 \\
XGBoost & 98.85 & 98.87 & 98.85 & 98.85 & 98.11 & 98.12 & 98.11 & 98.11 \\
Voting Classifier & 98.16 & 98.11 & 98.11 & 98.11 & 98.50 & 98.50 & 98.50 & 98.50 \\
Stacking Classifier & 97.93 & 98.62 & 97.93 & 98.19 & 98.60 & 98.60 & 98.60 & 98.60 \\
\midrule
\multicolumn{9}{c}{\textit{Feature Selection Methods (60 features)}} \\
\midrule
DT + RF Importance & 94.11 & 94.02 & 94.11 & 94.02 & 96.79 & 96.79 & 96.79 & 96.79 \\
RF + RF Importance & 97.88 & 97.56 & 97.88 & 97.71 & 98.00 & 98.01 & 98.00 & 98.00 \\
LR + RF Importance & 94.34 & 95.98 & 94.34 & 94.90 & 93.10 & 93.45 & 93.10 & 93.13 \\
\midrule
DT + Mutual Information & 91.58 & 91.54 & 91.58 & 91.48 & 96.16 & 96.16 & 96.16 & 96.16 \\
RF + Mutual Information & 95.72 & 95.44 & 95.72 & 95.56 & 97.60 & 97.61 & 97.60 & 97.61 \\
LR + Mutual Information & 86.20 & 88.57 & 86.20 & 86.71 & 85.49 & 86.30 & 85.49 & 85.56 \\
\midrule
DT + Chi-Squared & 92.18 & 92.31 & 92.18 & 92.08 & 83.69 & 85.06 & 83.69 & 83.78 \\
RF + Chi-Squared & 95.86 & 95.65 & 95.86 & 95.71 & 88.63 & 89.12 & 88.63 & 88.68 \\
LR + Chi-Squared & 92.46 & 93.91 & 92.46 & 92.90 & 89.06 & 90.09 & 89.06 & 89.12 \\
\midrule
DT + RFE & 95.77 & 95.48 & 95.77 & 95.57 & 97.04 & 97.04 & 97.04 & 97.04 \\
RF + RFE & 98.07 & 97.74 & 98.07 & 97.89 & 97.93 & 97.95 & 97.93 & 97.93 \\
LR + RFE & 94.53 & 95.88 & 94.53 & 94.98 & 94.26 & 94.51 & 94.26 & 94.28 \\
\midrule
\multicolumn{9}{c}{\textit{Proposed RL Methods having episodic length $\sim$60}} \\
\midrule
DDQN (Baseline) & \underline{99.12} & \underline{99.14} & \underline{99.13} & \underline{99.13} & \underline{98.83} & \underline{98.49} & \textbf{98.77} & \underline{98.63} \\
\textbf{D3QN (Proposed)} & \textbf{99.22} & \textbf{99.23} & \textbf{99.22} & \textbf{99.20} & \textbf{98.83} & \textbf{98.64} & \underline{98.65} & \textbf{98.64} \\
\bottomrule
\end{tabular}%
}
\begin{tablenotes}
\small
\item All metrics reported as percentages (\%)
\item \textbf{Bold}: Best overall performance per dataset; \underline{Underlined}: Second-best performing method per dataset
\item DT = Decision Tree, RF = Random Forest, LR = Logistic Regression, RFE = Recursive Feature Elimination
\end{tablenotes}
\end{table*}

\clearpage



\end{document}